\pdfoutput=1
\documentclass[11pt]{article}

\usepackage[preprint]{acl}
\usepackage{microtype}
\usepackage{arydshln}

\usepackage{times}
\usepackage{graphicx}
\usepackage{booktabs}
\usepackage{hyperref}
\usepackage{multirow}

\title{SinhalaMMLU: A Comprehensive Benchmark for Evaluating Multitask Language Understanding in Sinhala}

\author{
Ashmari Pramodya\textsuperscript{1}, 
Nirasha Nelki\textsuperscript{2}, 
Heshan Shalinda\textsuperscript{3}, 
Chamila Liyanage\textsuperscript{2}, 
Yusuke Sakai\textsuperscript{1}, \\
\textbf{Randil Pushpananda\textsuperscript{2}, Ruvan Weerasinghe\textsuperscript{3}, Hidetaka Kamigaito\textsuperscript{1}, Taro Watanabe\textsuperscript{1}} 
\\
\textsuperscript{1}Nara Institute of Science and Technology (NAIST), Japan \\
\textsuperscript{2}University of Colombo School of Computing (UCSC), Sri Lanka \\
\textsuperscript{3}Informatics Institute of Technology (IIT), Sri Lanka \\
\textit{\{pussewala.ashmari.ow4, sakai.yusuke.sr9, kamigaito.h, taro\}@is.naist.jp} \\
\textit{\{2021cs127@stu, rpn@, cml@\}ucsc.cmb.ac.lk, } 
\textit{\{heshan.20230591,ruvan.w\}@iit.ac.lk}
}

\begin{document}
\maketitle
\begin{abstract}

Large Language Models (LLMs) demonstrate impressive general knowledge and reasoning abilities, yet their evaluation has predominantly focused on global or anglocentric subjects, often neglecting low-resource languages and culturally specific content.
While recent multilingual benchmarks attempt to bridge this gap, many rely on automatic translation, which can introduce errors and misrepresent the original cultural context.
To address this, we introduce SinhalaMMLU, the first multiple-choice question answering benchmark designed specifically for Sinhala, a low-resource language.
The dataset includes over 7,000 questions spanning secondary to collegiate education levels and is aligned with the Sri Lankan national curriculum. It covers six domains and 30 subjects, encompassing both general academic topics and culturally grounded knowledge.
We evaluate 26 LLMs on SinhalaMMLU and observe that, while Claude 3.5 sonnet and GPT-4o achieve the highest average accuracies at 67\% and 62\% respectively, overall model performance remains limited. 
In particular, models struggle in culturally rich domains such as the Humanities, revealing substantial room for improvement in adapting LLMs to low-resource and culturally specific contexts.
\end{abstract}

\section{Introduction}
\label{sec:Intro}

The introduction of large language models (LLMs) has led to unprecedented improvements in natural language processing (NLP) and artificial intelligence, reshaping the field with their revolutionary capabilities \cite{openai2024gpt4technicalreport,anthropic2024claude3,touvron2023llamaopenefficientfoundation}.
Despite multilingual capabilities in LLMs, significant performance gaps persist between English and other languages \cite{DBLP:journals/corr/abs-2402-13524}.
This performance gap is particularly pronounced for languages using non-Latin scripts such as Sinhala.
One critical factor is the lack of high quality benchmarks for non-English languages, particularly those that are low resourced.
The majority of current evaluation frameworks remain English-centric \cite{hendryckstest2021}, creating significant assessment gaps for linguistically diverse and low-resource languages.
Many benchmarks claiming multilingual support are simply translated versions \cite{bandarkar-etal-2024-belebele,singh-etal-2025-global} of content originally designed for English speakers. 
These often include content that assumes knowledge of the American legal system or requires familiarity with English-specific cultural references and colloquialisms.
Despite appearing multilingual through translation, these benchmarks fail to reflect the cultural context  \cite{Ji_Ji_Bouillon_Seligman_2023} valued by native speakers.

\begin{figure}[t]
\centering{
\includegraphics[width=\linewidth]{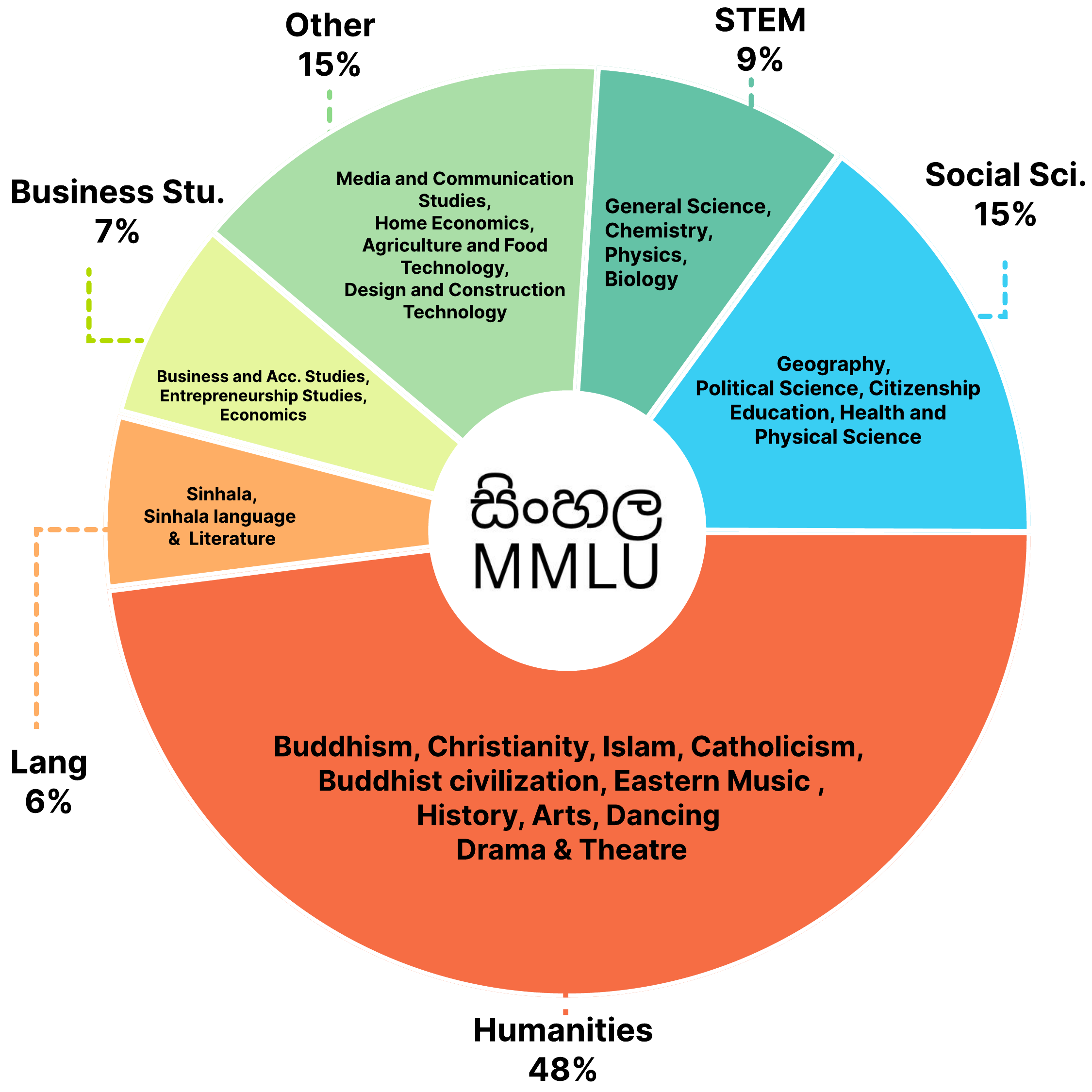}}
\caption{Distribution of the number of questions across different domains in SinhalaMMLU.}
\label{overview}
\end{figure}

To address the lack of existing benchmarks for the Sinhala NLP community and to better assess how LLMs perform on Sinhala, we introduce SinhalaMMLU, a comprehensive evaluation framework comprising over 7,000 questions across 6 domains and 30 subjects reflecting both general and culturally specific knowledge.
SinhalaMMLU is designed by collecting questions from various national and provincial exams, without relying on translations.
These questions include culturally relevant subjects such as Sri Lankan history, arts, drama and theatre, indigenous dancing, oriental music, and Sinhala language, alongside traditional academic subjects like science.

We evaluate 26 different LLMs, a mix of closed proprietary and open-source.
Our results show significant room for improvement, with Claude-sonnet 3.5 scoring the highest at 67.65\%.
We also explore how performance scales with the number of parameters, finding significant improvements as model size increases. 
Our domain-wise analysis reveals that models underperform in culturally relevant areas, such as Humanities and Language, as well as in general subjects within the STEM domain.
All data set details are publicly available on GitHub\footnote{\url{https://github.com/naist-nlp/SinhalaMMLU}}
and Hugging Face \footnote{\url{https://huggingface.co/datasets/naist-nlp/SinhalaMMLU}}.


\section{Related work}

\subsection{Benchmarks for LLMs}

In an era where different LLMs are emerging day by day, benchmarks are crucial to understanding their power and performance. Benchmarks like GLUE \cite{wang-etal-2018-glue}, SuperGLUE \cite{NEURIPS2019_4496bf24}, and SQuAD \cite{rajpurkar-etal-2016-squad} are being used as evaluation frameworks for natural language understanding (NLU) tasks, including reading comprehension and question-answering. Eventually, multilingual evaluation frameworks emerged like XGLUE \cite{liang-etal-2020-xglue} and XTREAM-R \cite{ruder-etal-2021-xtreme}. However, these benchmarks only assess the linguistic skills of the language rather than language understanding. 

MMLU \cite{hendryckstest2021} is one of the prominently used benchmarks for LLMs. It evaluates the LLMs via multiple-choice questions belonging to different subjects and domains to evaluate the knowledge acquisition of LLMs. Since MMLU consists of English questions, there are some attempts to create similar benchmarks for other languages, including Arabic \cite{koto-etal-2024-arabicmmlu}, Chinese \cite{li-etal-2024-cmmlu}, Turkish \cite{yuksel-etal-2024-turkishmmlu}, Indonesian \cite{koto-etal-2023-large}, Korean \cite{son-etal-2025-kmmlu}, and Persian \cite{Ghahroodi2024KhayyamC}. Some MMLU benchmarks have been constructed for low-resourced languages, notably African \cite{adelani-etal-2025-irokobench}, Malay \cite{poh-etal-2024-malaymmlu}, and Basque \cite{DBLP:conf/nips/EtxanizASLA24}. The MILU benchmark that consists of 11 Indic languages, stands out among the language-specific MMLU \citep{verma-etal-2025-milu}. In an effort to establish a multilingual benchmark, the Include benchmark \cite{romanou2024includeevaluatingmultilinguallanguage} has achieved a remarkable feat as a benchmark by creating an impressive QA dataset that encompasses a variety of exams from various languages. 

\subsection{Sinhala Language}

Sinhala is an Indo-Aryan language spoken by over 17 million people in Sri Lanka, where it serves as one of the two official languages. 
Despite its sizable community, Sinhala remains relatively under-resourced in terms of language technology and NLP tools, though there have been some attempts for it.

The existing textual datasets for Sinhala have been curated for different purposes like text summarization \cite{hasan-etal-2021-xl, 10.1007/978-3-031-70248-8_17}, and text classification \cite{hettiarachchi-etal-2024-nsina, warusawithana-etal-2022-systematic, Ranasinghe2024}.
The Aya Dataset \cite{singh-etal-2024-aya}, a human-annotated multilingual instruction fine-tuning dataset, also includes Sinhala data. However, these works have only focused on a single task or a single domain; therefore, not varied enough to consider for a multi-task benchmark for LLMs.
Since the MMLU has been proposed for English, there have been attempts to create multilingual benchmarks by translating the initial dataset into other languages, including Sinhala \citep{singh-etal-2025-global}. However, there is a question regarding the accuracy of the translation, as it can be challenging to translate language-specific terms. A language-specific benchmark or a multilingual benchmark should consider culture-related aspects as well, in order to ensure LLMs understand the cultural contexts that are lacking in the GlobalMMLU \citep{singh-etal-2025-global}.


\section{Sinhala MMLU}
Motivated by the scarcity of high-quality resources for Sinhala, we introduce SinhalaMMLU, a benchmark designed to evaluate language models in the Sinhala language.
The benchmark includes multiple-choice questions contextually designed for Sinhala language, ranging over a wide range of subjects and educational levels. 
Following the structure and methodology of the original English MMLU benchmark, we carefully curated the dataset in alignment with the local national curriculum to ensure both relevance and cultural appropriateness.

The national educational curriculum of Sri Lanka is divided into three broad levels: primary, secondary, and collegiate education. Primary education spans five years (Grades 1-5) for children aged 5 to 9, focusing on foundational subjects. However, we excluded this level from SinhalaMMLU, as the questions are generally simple (e.g., drawing, filling blanks, or writing), and are not suited for multiple-choice evaluation. Secondary education is divided into two phases: junior secondary (Grades 6 to 9, ages 10 to 13) and senior secondary (Grades 10 to 11, ages 14 to 16), culminating in the General Certificate of Education Ordinary Level (GCE O-Level) examination, where students take six main subjects and three additional subjects. The collegiate level comprises Grades 12-13 (ages 17-19) and concludes with the GCE Advanced Level (A-Level) examination. There are five academic streams at this level: Arts, Commerce, Biological Science, Physical Science, and Technology. Students select at least three subjects within their chosen stream, and the national examination includes these core subjects along with a General English test and a Common General Test. The A-Level exam serves as the primary qualification for university entrance in Sri Lanka.
\begin{table}[t]
\resizebox{\columnwidth}{!}{%
\centering
\begin{tabular}{@{}lrcr@{}}
\toprule
\textbf{Group} & \textbf{\# Questions} & \multicolumn{2}{c}{\textbf{\# Chars}} \\ \cmidrule(l{0.25em}r{0.25em}){3-4} 
 & & \textbf{Question} & \textbf{Answer} \\
\midrule
Easy       &  1893      &   59.08     &    16.77   \\
Medium        &  2585       &   100.66     &  24.79     \\
Hard          &  2566      &    116.40    &   27.53    \\ \midrule
STEM          &  629     &  157.82      & 27.42      \\
Social Science&  1084       &  141.80     & 22.34      \\
Humanities    &  3419      &   93.91     &  22.24     \\
Language      &  397      &  74.19      &   25.65    \\
Business Studies & 477       &  173.39      & 32.99       \\
Other         &  1038       &  108.58      &   28.24    \\ 
\bottomrule
\end{tabular}%
}
\caption{Total number of questions, Average question and answer length (in characters) for each difficulty level and domain.The overall question count is 7044.}
\label{tab:data_stat}

\end{table}

\begin{table}[ht]
\centering
\resizebox{\columnwidth}{!}{%

\begin{tabular}{ll}
\toprule
\textbf{Domain} & \textbf{Subject} \\
\midrule
Humanities & History \\
           & Drama and Theatre \\
           & Dancing \\
           & Eastern Music \\
           & Arts \\
           & Buddhism \\
           & Catholicism \\
           & Christianity \\
           & Islam \\
           & Buddhist Civilization \\
           & Oriental Music \\
           & History of Sri Lanka \\
           & Dancing Indigenous \\
\midrule
Social Science & Citizenship Education \\
               & Health and Physical Science \\
               & Geography \\
               & Political Science \\
\midrule
STEM & Physics \\
     & Chemistry \\
     & Biology \\
     & Science \\
\midrule
Language & Sinhala Language and literature \\
\midrule
Business Studies & Business and Accounting Studies \\
                 & Entrepreneurship Studies \\
                 & Economics \\
\midrule
Other & Home Economics \\
      & Biosystems Technology \\
      & Communication and Media Studies \\
      & Design and Construction Technology \\
      & Agriculture and Food Technology \\
\bottomrule
\end{tabular}%
}
\caption{Subjects categorized by domain in the SinhalaMMLU dataset.}
\label{tab:subjects_by_domain}
\end{table}

All questions were prepared in alignment with the official curriculum set by the Ministry of Education (MOE), Sri Lanka. 
By aligning SinhalaMMLU with educational standards, we aim to provide a rigorous and contextually grounded benchmark to evaluate the performance of LLMs in Sinhala across educational levels. 
This allows us to systematically assess model understanding of local knowledge, identify linguistic or subject-specific weaknesses, and encourage the development of culturally relevant and educationally aligned language models for low-resource languages.


\subsection{Data Preparation}

Our primary data source was e-thaksalawa\footnote{\url{https://www.ethaksalawa.moe.gov.lk/En/index.php}}, the public official e-learning platform developed by the Sri Lankan MOE.
This platform provides free access to a wide range of educational materials, including past exam papers, lecture notes, interactive textbooks, video lessons, and multiple-choice questions (MCQs) with answer keys curated by domain experts.
We collected MCQs aligned with the national curriculum from government-focused papers, specifically from the GCE O-Level, GCE A-Level, and provincial examination papers for Grades 6–8. 
These exams include multiple-choice sections covering various academic subjects. In addition to e-Thaksalawa, we used other educational websites that host government-released past papers and marking schemes.

To curate the data, we employed four annotators with undergraduate or higher educational qualifications. 
Over a two-month period, the annotators manually extracted MCQs from PDF documents (after OCR), while quiz-based content was scraped directly from the web platform\footnote{\url{https://www.e-thaksalawa.moe.gov.lk/moodle/course/view.php?id=636}}. Annotators were instructed to include only questions that were in multiple-choice format with clearly defined answer options and an identified correct answer.  
Questions containing multimodal content, such as images, audio, or video, were excluded.
Furthermore, we excluded Mathematics questions because they primarily involve symbolic reasoning and are not usually provided as MCQ questions.
More than 65\% of the collected data was sourced from PDFs.

\begin{figure*}[t]
\centering{
\includegraphics[width=\linewidth]{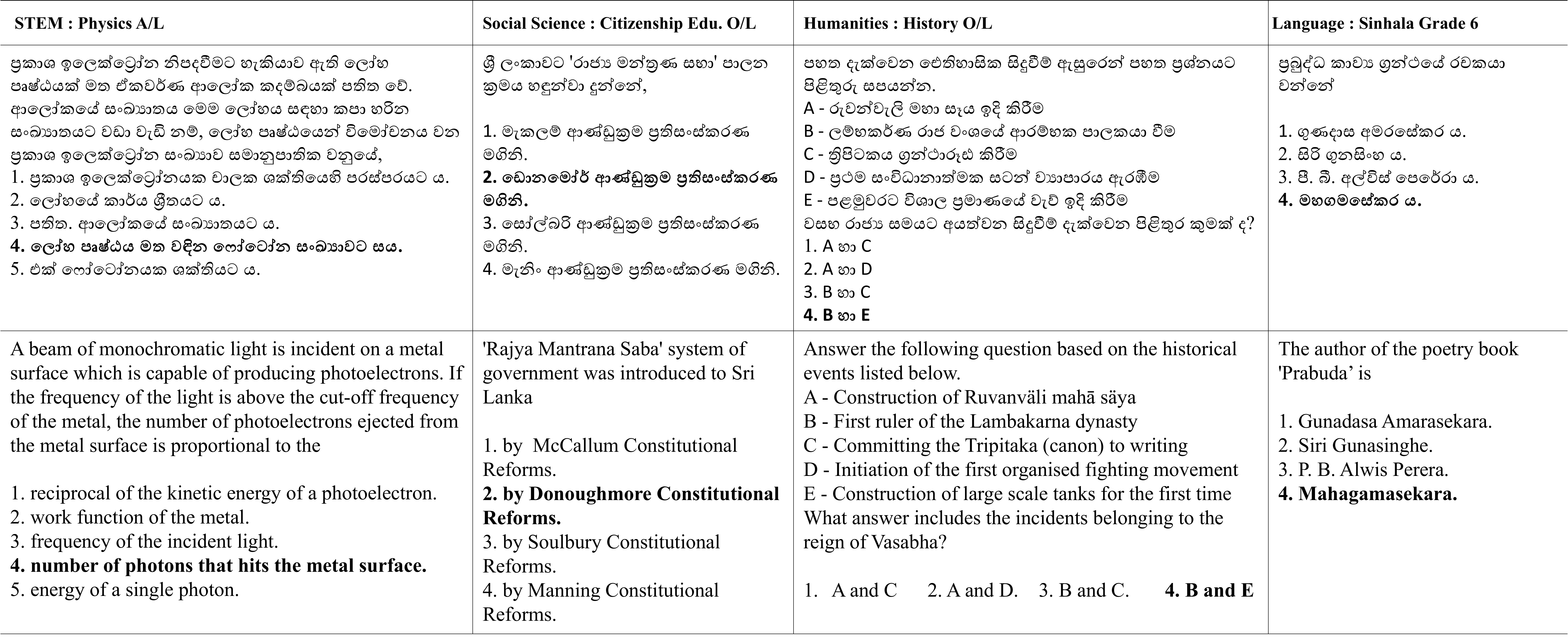}}
\caption{Examples of questions from 4 domains with answers in \textbf{bold}. English translations are provided below each question. The table also includes a sample suboption-type question from the Humanities domain.}
\label{example_questions}
\end{figure*}

\subsection{Data Formatting}

Each MCQ in SinhalaMMLU follows a consistent structure composed of a question stem, a set of four or five choices, and a single correct answer. The dataset includes both fill-in-the-blank style questions and direct queries. Along with each question, we recorded metadata such as the subject, difficulty level mapped to school grade, original source and year, and the province, especially in the case of government-issued or provincial papers. 
For chemical formulae and mathematical expressions, we use a 50:50 mixture of LATEX and plain text, where plain text was only allowed if an expression is commonly used and not prone to ambiguity.
To ensure content uniqueness and eliminate redundancy,  we removed exact duplicate questions with identical answer sets and applied cosine similarity analysis. Questions with a similarity score exceeding 95\% were flagged and removed from the dataset.
Example questions are illustrated in Figure~\ref{example_questions}.

\subsection{Difficulty Classification}
\label{sec:difficulty_classify}

We classified question difficulty as follows.
Questions from Grades 6–8 were labeled as \textbf{easy}. 
Most of these had four answer choices, although a subset of geography questions (44 in total) included only three. 
To maintain format consistency, we added a fourth option using Claude 3.7 Sonnet, which was not used during model evaluation with our dataset.
Questions from Grades 9–11 were considered \textbf{medium} in difficulty, all of which contained four answer options. 
Questions from Grades 12–13 were labeled as \textbf{hard} and typically featured five answer choices. 
This classification mirrors the increasing academic depth and complexity expected at higher grade levels and provides a clear basis for evaluating model performance across educational stages.

\subsection{Data Distribution}

Table \ref{tab:data_stat} summarizes the dataset statistics, including the average question length across 7,044 questions.
Each subject has at least 104 questions, which we split into few-shot set with 3 questions, and test set with more than 100 questions.
As shown in Table~\ref{tab:data_stat} and accompanying Figure~\ref{overview}, our SinhalaMMLU includes the following domains: ``Humanities'' which is deeply rooted in Sri Lankan cultural and historical context; 
``STEM'',``Social Science'',``Business Studies'', ``Language'', and ``Other'', 
For a detailed breakdown of the subjects within each domain and their difficulty levels in Table~\ref{tab:subjects_by_domain}, refer to Appendix~\ref{sec:app:subjects}.


\section{Experimental Setup }

\paragraph{Models} We evaluated 26 recent state-of-the-art multilingual LLMs of different sizes in zero-shot and few-shot settings. 
These include both small and large open models as well as closed models. 
For open source models, we include Cohere4AI's Aya Expanse\footnote{\url{https://hf.co/blog/aya-expanse}}, LLaMA-3 \cite{touvron2023llamaopenefficientfoundation}, Qwen \cite{qwen2025qwen25technicalreport}, and Mistral \cite{jiang2024mixtralexperts}, along with their chat versions.
To optimize memory efficiency during inference, we applied 4-bit NF4 quantization with double quantization and bfloat16 computation. 
For closed models, we used GPT-4o \cite{openai2024gpt4technicalreport} and Claude \cite{anthropic2024claude3}. The details of these models are provided in the Appendix~\ref{sec:Model Details}.

\paragraph{Evaluation} We evaluated LLMs by accuracy. Following previous studies \cite{koto-etal-2023-large,li-etal-2024-cmmlu}, for open-source models, we determine the answer based on the highest probability among all possible options.  
For closed-source models, we determine the answer based on the first token generated in the text using a regular expression.


\begin{table*}[t]
\centering
\resizebox{\textwidth}{!}{%
\begin{tabular}{@{}lccccccc@{}}
\toprule
{Model} &{Humanities} & {Language} & {Social Science} & {STEM} & {Business Studies} & {Other} & {Average} \\
\midrule
\textsc{Claude-3-5-sonnet} & \textbf{66.15}\ & \textbf{62.37}\ & \textbf{77.55}\ & \textbf{61.40}\ & \textbf{73.12}\ & \textbf{65.58}\ & \textbf{67.65}\ \\
\textsc{Claude-3-haiku} & 41.01\ & 43.81\ & 50.57\ & 35.34\ & 44.09\ & 39.64\ & 42.14\ \\ \hdashline

\textsc{GPT-4o} & 62.02\ & 51.29\ & 71.32\ & 60.59\ & 67.53\ & 61.44\ & 62.95\ \\ \midrule
\textsc{Qwen2.5-7B} & 22.27\ & 21.65\ & 25.66\ & 18.57\ & 20.22\ & 21.70\ & 22.20\ \\
\textsc{Qwen2.5-7B-chat} & 28.02\ & 25.00\ & 28.30\ & 28.50\ & 27.10\ & 23.67\ & 27.23\ \\
\textsc{Qwen2.5-32B} & 28.85\ & 28.35\ & 35.28\ & 26.38\ & 25.16\ & 27.51\ & 29.15\ \\
\textsc{Qwen2.5-32B-chat} & 36.70\ & 32.73\ & 38.96\ & 38.60\ & 34.84\ & 34.02\ & 36.47\ \\
\textsc{Qwen2.5-72B} & 34.69\ & 28.87\ & 40.57\ & 37.95\ & 28.82\ & 34.52\ & 35.14\ \\
\textsc{Qwen2.5-72B-chat} & 39.84\ & 38.92\ & 45.85\ & 45.60\ & 41.29\ & 38.86\ & 41.18\ \\  \hdashline
\textsc{Llama-3.2-1B} & 22.18\ & 21.65\ & 25.66\ & 18.40\ & 19.78\ & 21.70\ & 22.12\ \\
\textsc{Llama-3.2-1B-chat} & 22.27\ & 21.65\ & 25.75\ & 18.73\ & 19.78\ & 21.70\ & 22.20\ \\
\textsc{Llama-3.2-3B} & 22.18\ & 22.68\ & 25.57\ & 18.40\ & 19.57\ & 21.70\ & 22.14\ \\
\textsc{Llama-3.2-3B-chat} & 22.27\ & 21.65\ & 25.66\ & 18.57\ & 19.35\ & 21.70\ & 22.14\ \\

\textsc{Llama-3-8B} & 22.48\ & 21.65\ & 26.32\ & 19.54\ & 20.86\ & 21.50\ & 22.51\ \\
\textsc{Llama-3-8B-chat} & 22.87\ & 21.65\ & 26.98\ & 19.38\ & 19.78\ & 23.18\ & 22.96\ \\
\textsc{Llama-3-70B} & 22.48\ & 23.45\ & 24.95\ & 22.48\ & 19.78\ & 22.09\ & 22.65\ \\
\textsc{Llama-3-70B-chat} & 27.15\ & 20.36\ & 26.32\ & 23.78\ & 20.65\ & 22.49\ & 25.21\ \\
\textsc{Llama-3.1-8B} & 22.90\ & 23.45\ & 27.17\ & 19.54\ & 23.44\ & 23.37\ & 23.39\ \\
\textsc{Llama-3.1-8B-chat} & 25.29\ & 25.52\ & 29.06\ & 22.31\ & 21.08\ & 24.95\ & 25.28\ \\
\textsc{Llama-3.1-70B}& 22.19\ & 22.94\ & 26.42\ & 19.54\ & 19.78\ & 21.60\ & 22.40\ \\
\textsc{Llama-3.1-70B-chat} & 27.24\ & 26.80\ & 25.94\ & 23.78\ & 26.67\ & 25.25\ & 26.37\ \\ 
\textsc{Llama-3.3-70B-chat} & 24.54\ & 23.45\ & 27.26\ & 23.13\ & 25.81\ & 22.88\ & 24.61\ \\ \hdashline
\textsc{Mistral-7B-chat} & 21.55\ & 22.94\ & 26.92\ & 22.36\ & 20.00\ & 20.81\ & 22.28\ \\
\textsc{Mistral-7B} & 22.18\ & 21.65\ & 25.66\ & 18.40\ & 19.78\ & 21.65\ & 22.12\ \\ \hdashline
\textsc{aya-expanse-8b} & 22.78\ & 24.23\ & 25.09\ & 19.71\ & 20.43\ & 21.70\ & 22.62\ \\
\textsc{aya-expanse-32b} & 23.97\ & 25.00\ & 30.00\ & 25.08\ & 23.66\ & 24.65\ & 25.14\ \\ 
 \hdashline
Avg & 27.37 & 26.10 & 31.79 & 27.16 & 27.78 &  28.07 & 28.38  \\ 
\bottomrule
\end{tabular}%
}
\caption{Zero-shot performance (\% accuracy) of LLMs across the six domains.}
\label{tab:zero-shot-structure}
\end{table*}

\paragraph{Prompt} For the SinhalaMMLU evaluation, we used two distinct prompt templates, one that specifies the subject domain and another that omits this information. 
Both follow the instruction format proposed by \citet{hendryckstest2021}, with prompt instructions given in the same language as the question.
More details in Appendix~\ref{sec:prompt}.


\section{Results and Discussions}

This section presents the evaluation results on the SinhalaMMLU dataset. To assess the impact of subject-specific information in prompts, we compared two variants: one incorporating the subject name and one without.
We observed that including the subject name generally improves performance across models. Therefore, for all subsequent experiments, we report results using prompts that include subject information.
Appendix~\ref{sec:prompt_compare} provides a detailed comparison. 


\paragraph{Results  by model}

\begin{figure}[t]
\centering{
\includegraphics[width=\linewidth]{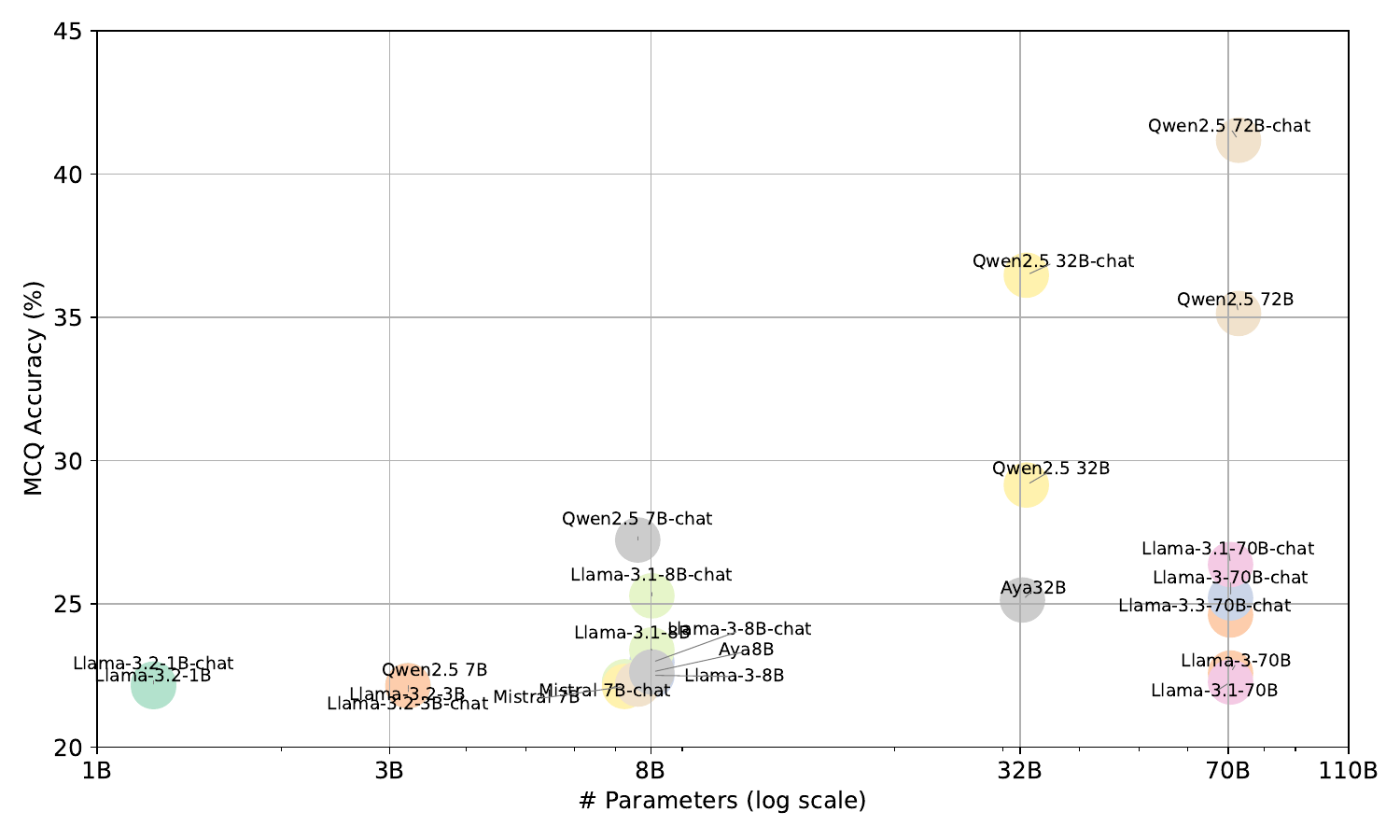}}
\caption{Individual models plotted by parameter count (x-axis, log scale) against multiple-choice question (MCQ) accuracy on the SinhalaMMLU benchmark (y-axis).}
\label{model_scaling}
\end{figure}

Table~\ref{tab:zero-shot-structure} summarizes the full results of all the models, grouped by domain.
Closed models, notably Claude 3.5 Sonnet and GPT-4o, exhibit superior performance across all domains, achieving average accuracies of 67.65\% and 62.95\%, respectively. These models consistently outperform open-source counterparts, indicating a higher level of generalization and understanding.
Among open-source models, Qwen2.5-72B-chat and Qwen2.5-32B-chat stand out, with average accuracies of 41.18\% and 36.47\%, respectively. Their performance suggests that larger model sizes and instruction tuning contribute positively to handling diverse and complex tasks. 
While within the LLaMA-3 model family, LLaMA-3.1-70B-chat is the only variant that achieves comparatively higher performance.
In contrast, smaller models like LLaMA-3.2-1B and Mistral-7B variants show lower performance, with average accuracies hovering around 22\%. This disparity underscores the importance of model scale and training data diversity in achieving higher accuracy.
Figure~\ref{model_scaling} illustrates that the scaling law is still present, as it is evident that models with large sizes perform better than smaller ones.
A similar trend across model families is also observed in \citet{winata-etal-2025-worldcuisines}.

\paragraph{Results by Domain}

Models consistently underperformed in culturally grounded areas, with Humanities and Language achieving average scores of 66.15\% and 62.37\% respectively with Claude. These results reflect the inherent challenges these models face when navigating cultural context and idiomatic expressions that are deeply embedded in local knowledge. Similarly, performance in STEM was also limited, with an average of 61.40\%, likely due to the difficulty of handling domain-specific terminology and complex conceptual reasoning. However, Social Sciences demonstrated markedly stronger performance, achieving 77.55\% accuracy. This success can be attributed to the predominantly factual and descriptive nature of subjects such as Geography, Citizenship Education, and Health, which rely on standardized terminology that is well represented in the training data of most multilingual LLMs.
Appendix~\ref{sec:Subject_results} provides more details.

\paragraph{Results across different difficulty levels}

\begin{figure}[t]
\centering{
\includegraphics[width=\linewidth]{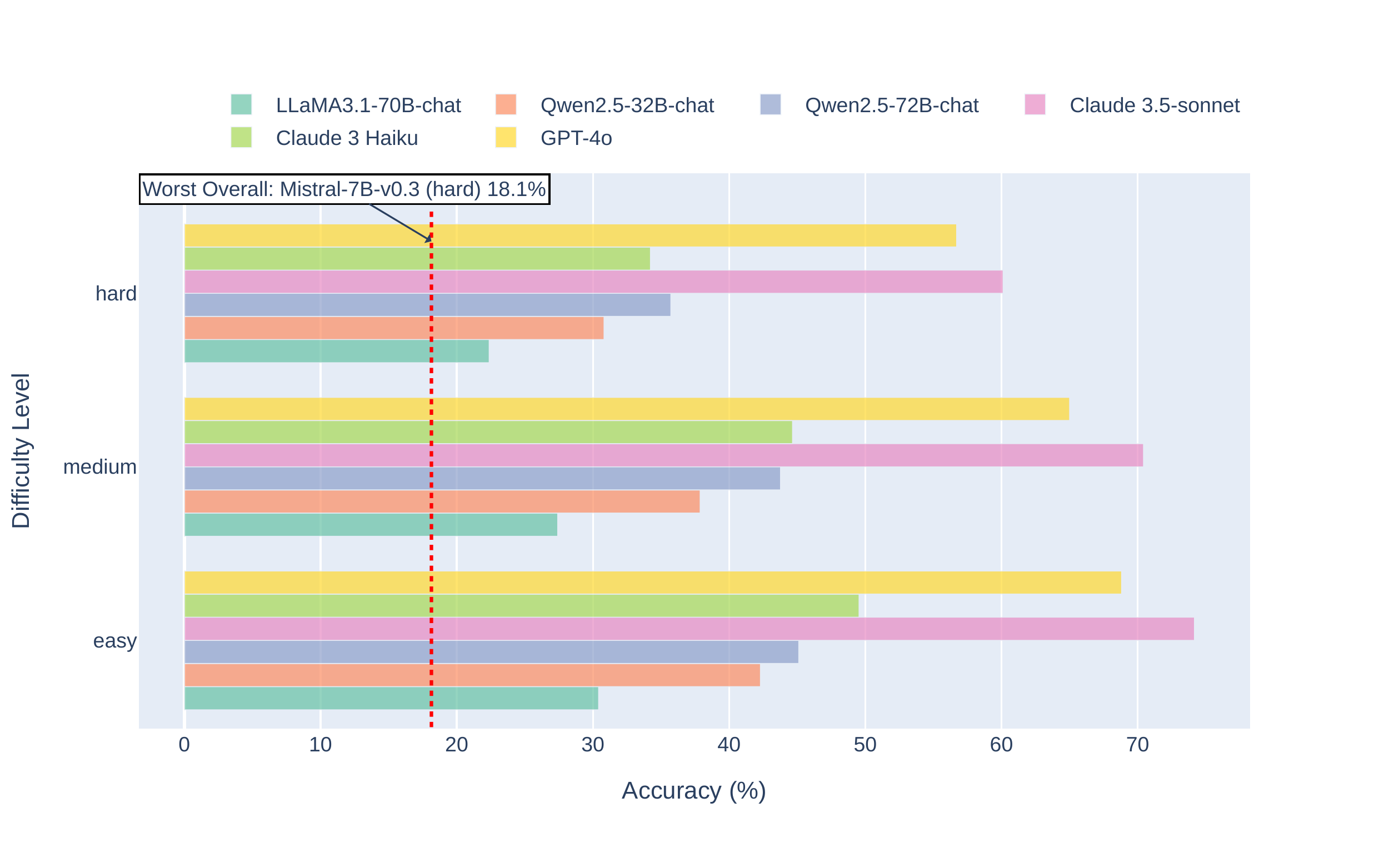}}
\caption{LLM performance across different difficulty levels.}
\label{diifcultBarplot}
\end{figure}

Figure~\ref{diifcultBarplot} presents the average scores of the top-performing models (Claude 3.5 Sonnet, GPT-4o, Claude 3 Haiku, Qwen-72B-Chat, Qwen-32B-Chat, and LLaMA-3.1-70B-Chat) across the difficulty levels defined in Section~\ref{sec:difficulty_classify}. 
The results indicate that Sinhala questions at the collegiate level (hard), which typically include five answer options, pose greater challenges for all models.


\begin{table}[t]
\centering
\resizebox{\columnwidth}{!}{%
\begin{tabular}{lcc}
\toprule
{Model} & {0-shot (\%)} & {3-shot (\%)} \\
\midrule
\textsc{Qwen2.5-72B-Chat} & 41.18 & 41.24  \\
\textsc{Qwen2.5-72B} & 35.14 & 35.20 \\
\textsc{Qwen2.5-32B-Chat} & 36.47 & 37.54 \\
\textsc{Qwen2.5-32B} & 29.15 & 29.68  \\
\textsc{LLaMA-3.1-70B-chat} & 26.37 & 25.05 \\
\textsc{LLaMA-3.1-70B} & 22.40 & 22.89 \\
\bottomrule
\end{tabular}%
}
\caption{Accuracy comparison of models under 0-shot and 3-shot settings}
\label{tab:zero_vs_few_shot}
\end{table}


\paragraph{Few shot vs zero shot}
As shown in Table~\ref{tab:zero_vs_few_shot}, few-shot prompting results in limited or inconsistent improvements across top-performing open-source models. While base models such as Qwen-32B show marginal gains, instruction-tuned models like LLaMA-3.1-70B-chat exhibit performance degradation, likely due to their optimization for dialogue-based tasks rather than in-context learning.
These findings are consistent with prior work by \citet{verma-etal-2025-milu}, which also observed similar limitations in instruction-tuned models under few-shot settings.
More detailed in Appendix \ref{sec: app_fewshot}.


\paragraph{Is negations challenging?}
Negation is frequently used in Sri Lankan school exam questions to increase question difficulty and evaluate students' capacity for reasoning. 
In the domain of NLP, prior work has similarly demonstrated that negation poses significant challenges to model performance \cite{truong-etal-2023-language}. 
To examine its impact within our benchmark, we adopt a simple string-matching method to identify and analyze questions containing negation.
We utilize specific negation phrases to identify questions containing negations in Sinhala.
These include: \includegraphics[height=1.5\fontcharht\font`\B,trim=0 1mm 0 -1mm]{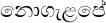} (Not compatible), \includegraphics[height=1.4\fontcharht\font`\B,trim=0 1mm 0 -1mm]{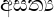} (False), \includegraphics[height=1.5\fontcharht\font`\B,trim=0 1mm 0 -1mm]{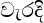} (wrong), \includegraphics[height=1.5\fontcharht\font`\B,trim=0 1mm 0 -1mm]{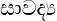} (False).
In Table~\ref{tab:negation-accuracy}, most models perform less effectively on questions containing negation compared to those without, including Claude 3.5 Sonnet. GPT-4o is the only model that maintains robust performance. This pattern aligns with previous findings reported by~\citet{li-etal-2024-cmmlu}.


\begin{table}[t]
\resizebox{\columnwidth}{!}{%
\centering
\begin{tabular}{lcccc}
\toprule
{Model} & \multicolumn{2}{c}{{Negations}} & \multicolumn{2}{c}{{Suboptions}} \\ \cmidrule(l{0.25em}r{0.25em}){2-3} \cmidrule(l{0.25em}r{0.25em}){4-5} 
               & {w/} & {w/o} & {w/} & {w/o} \\ 
\midrule
\textsc{claude-3-5-sonnet}& 58.36 & 67.66 & 57.45 & 68.72\\
\textsc{GPT-4o}            & 59.59 & 62.91 & 53.14 & 64.05 \\
\textsc{claude-3 Haiku}   & 32.65 & 42.35 & 28.11 & 43.60 \\
\midrule
\textsc{Qwen2.5-72B-Chat}    & 25.30 & 41.82 & 34.10 & 41.91 \\
\textsc{Qwen2.5-32B-Chat}    & 24.48 & 37.12 & 26.88 & 37.47 \\
\textsc{LLaMA-3.1-70B-chat}& 22.45 & 26.00 & 20.73 & 26.96\\
\bottomrule
\end{tabular}%
}
\caption{Average accuracy classified by questions with and without negation expressions and Suboptions.}
\label{tab:negation-accuracy}
\end{table}

\paragraph{Impact of Suboption Questions}
Similar to negation-based questions, suboption questions are another type used in exams to make questions more challenging.
These questions typically consist of a main question followed by multiple subparts (e.g., labeled A, B, C, D).
Common formats include selecting the correct sequence of events, matching items across suboptions, or determining whether one statement correctly explains another.
These questions account for about 9.8\% of the data.
They often demand models to engage in deeper reasoning, including multi-step inference, comparison of related statements, and understanding of logical or causal relationships. 
According to Table~\ref{tab:negation-accuracy}, we observed that all models have weaker performance on these questions.

\begin{table}[t]
\centering
\resizebox{\columnwidth}{!}{%
\begin{tabular}{lccc}
\toprule
{Model} & {5 options} & { 4 options} & {$\Delta$} \\
\midrule
\textsc{claude-3-5-sonnet}  & 60.11 & 66.13 & +6.02 \\
\textsc{GPT-4o}  & 56.71 & 62.17  & +5.46 \\
\textsc{claude-3 Haiku}  & 34.21 & 39.44 & +5.23 \\
\textsc{Qwen2.5-72B-Chat}     & 35.72 & 40.95  & +6.74 \\
\textsc{Qwen2.5-32B-Chat}   & 30.81 & 34.73 & +3.92 \\
\textsc{LLaMA-3.1-70B-chat} & 22.38  &  26.81  & +4.43\\
\bottomrule
\end{tabular}%
}
\caption{Performance comparison on the original SinhalaMMLU and its culturally grounded subset.}
\label{tab:hard}
\end{table}

\section{Do Hard Questions Get Easier with Fewer Options?}

In Sri Lankan GCE A-Level exams, which fall under the hard difficulty level in our classification, multiple choice questions typically include five answer choices, unlike the standard four used in most evaluation benchmarks.
To investigate whether reducing the number of options improves model performance, we compared accuracy between the original 5-option format and a modified 4-option version, in which one incorrect option was randomly removed.
As shown in Table~\ref{tab:hard}, all models improved with fewer options. Claude 3.5 Sonnet and GPT-4o saw gains of +6.02 and +5.46 points, respectively. 
However, even with 4 choices, performance remained lower than on medium-difficulty questions, suggesting that reduced options do not fully offset the complexity of hard-level content.
To further explore model behavior, we analyzed confidence scores across both settings. Interestingly, while confidence scores tend to be higher in the 4-option setup (Table~\ref{tab:conf-acc-corr}), correlation between confidence and accuracy remains weak or inconsistent across models. 
For instance, Qwen2.5-72B-chat shows moderate positive correlation in the five-option format. 
These findings suggest that reducing the number of answer choices can simplify the task for LLMs and lead to better performance, particularly on harder, culturally grounded questions. 
However, improved accuracy does not always align with model confidence.

\begin{table}[t]
\resizebox{\columnwidth}{!}{%

\centering
\begin{tabular}{lcc}
\toprule
{Model} & {4 Options (Corr)} & {5 Options (Corr)} \\
\midrule
\textsc{Qwen2.5-72B-chat} & $0.6835$ & $0.7588$ \\
\textsc{Qwen2.5-32B-chat} & $-0.0045$ & $0.0696$ \\
\textsc{LLaMA-3.1-70B-chat}& $-0.0602$ & $-0.5529$  \\
\bottomrule
\end{tabular}%
}
\caption{Correlation between model confidence and accuracy under 4-option and 5-option settings.}
\label{tab:conf-acc-corr}
\end{table}

\section{Do Translated STEM Questions Match the Quality of Native Ones?}

Translation is the simplest way to scale multilingual benchmarks. Attempts have been made to translate MMLU to Sinhala \cite{singh-etal-2025-global}. 
A key challenge in adapting those benchmarks to low-resource languages is the accurate translation of domain-specific terminology, particularly in scientific subjects. 
In Sinhala, scientific concepts are conveyed through precise and contextually appropriate vocabulary that reflects standardized usage in the national curriculum.
Literal or machine-translated versions from English frequently fail to capture these nuances, resulting in awkward or unnatural phrasing that deviates from actual usage.

To assess the linguistic quality of translated versus native content, we randomly sampled 100 questions from our native Sinhala MMLU benchmark and 100 questions from a Sinhala-translated subset of the global MMLU STEM dataset, excluding mathematics and engineering.
Two Sinhala-speaking annotators from the Biology stream\footnote{In the Sri Lankan Advanced Level (A/L) education system, students specialize in one of five streams: Biology, Physical Science, Commerce, Arts, or Technology. The Biology stream includes biology and chemistry as compulsory subjects, along with either physics or agriculture as a third subject. The annotators had studied biology, chemistry, and physics under this stream.} of Sri Lanka's Advanced Level (A/L) education system were asked to rate each question's linguistic naturalness on a 5-point scale, focusing solely on language fluency and terminology, independent of the correctness of the answer.

Table~\ref{tab:natural} shows the comparison of the naturalness. Native Sinhala content scored significantly higher in linguistic fluency (97.3) compared to translated questions (71.07).
Our analysis reveals that while translation offers a scalable approach, it often fails to capture the precise technical terms used in Sinhala STEM subjects, resulting in awkward or misleading phrasing. For instance, ``plasmolysis'' was incorrectly translated as \includegraphics[height=1.5\fontcharht\font`\B,trim=0 1mm 0 -1mm]{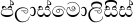} (plāsmolisis) instead of the proper technical term \includegraphics[height=1.5\fontcharht\font`\B,trim=0 1mm 0 -1mm]{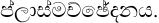} (plāsmacchēdanaya).
Together, these results emphasize that direct translation is insufficient for creating high-quality multilingual benchmarks \cite{sakai-etal-2024-mcsqa}.

\begin{table}[t]
\resizebox{\columnwidth}{!}{%
\centering
\begin{tabular}{lc}
\toprule
{Data set} & {Score (\%)} \\
\midrule
SinhalaMMLU (Ours) & 97.30 \\
GlobalMMLU-si \cite{singh-etal-2025-global} & 71.07 \\
\bottomrule
\end{tabular}%
}
\caption{Naturalness ratings: GlobalMMLU-si vs. SinhalaMMLU STEM questions. We applied a linear transformation to convert the 5-point naturalness scores into a 100-point scale for ease of interpretation.}
\label{tab:natural}
\end{table}

\section{Cultural Knowledge Analysis}

To analyze how well LLMs understand Sri Lankan culture and the linguistic characteristics of the Sinhala language, we manually annotated questions that require knowledge of Sinhala vocabulary, literature, and culturally embedded concepts.
Most culturally relevant items were concentrated in domains such as drama and theatre, oriental music, traditional dance, Sinhala language and literature, and Sri Lankan history. 

\begin{figure}[t]
\centering{
\includegraphics[width=0.7\linewidth]{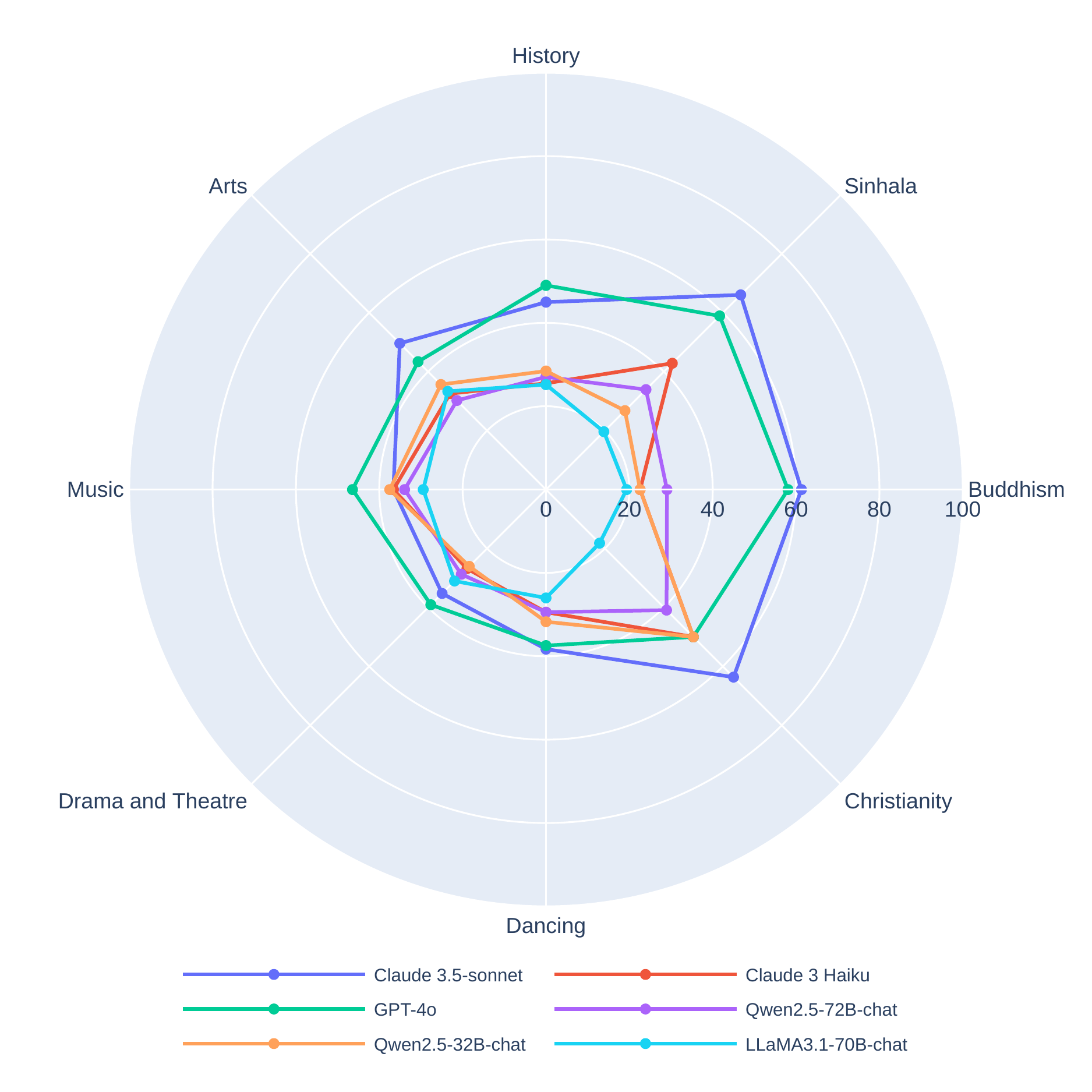}}
\caption{Subject-wise accuracy of LLMs on a culturally significant subset of the SinhalaMMLU dataset.}
\label{culture_radar}
\end{figure}

In this section, we present the performance of selected models on these culturally grounded questions.
The SinhalaMMLU cultural subset consists of a collection of 1,608 (22\%) handpicked questions.
Table~\ref{tab:Sinhalmmlu-cul} presents a performance analysis, focusing
on the discrepancy between general performance and Sinhala cultural-specific questions, revealing a consistent performance drop across all models on this cultural subset.
Closed-source models such as Claude 3.5 Sonnet and GPT-4o, experienced substantial drops of 28.22 and 23.83 percentage points when evaluated solely on culturally relevant questions.
Interestingly, LLaMA-3.1-70B-Chat showed the smallest drop (-0.01), but its original performance was already significantly lower (26.37\%), indicating limited overall capability.
Our subject-wise analysis (Figure~\ref{culture_radar}) reveals 
Claude 3.5 Sonnet performs particularly well in areas related to the Sinhala language and Buddhist concepts, while exhibiting weaknesses in domains such as traditional music, drama, and history. 
In contrast, GPT-4o demonstrates complementary strengths, with comparatively better performance in subjects related to music and historical knowledge.

\begin{table}[t]
\centering
\resizebox{\columnwidth}{!}{%
\begin{tabular}{lccc}
\toprule
{Model} & {Original} & { Cul.} & {$\Delta$} \\
\midrule
\textsc{claude-3-5-sonnet}  & 67.65 & 39.43 & -28.22 \\
\textsc{GPT4o}  & 62.95 & 39.12 & -23.83 \\
\textsc{claude-3 Haiku}  & 42.14 & 28.11  & -14.03 \\
\textsc{Qwen-72B-Chat}     & 41.18 & 30.03 & -11.15 \\
\textsc{Qwen-32B-Chat}   & 36.17 & 28.36 & -7.81 \\
\textsc{LLaMA-3.1-70B-chat} & 26.37 & 26.36  & -0.01 \\
\bottomrule
\end{tabular}%
}
\caption{Performance comparison on the original SinhalaMMLU and and its culturally grounded subset(Cul).}
\label{tab:Sinhalmmlu-cul}
\end{table}

\paragraph{Error Analysis} GPT-4o correctly answered 241 questions that Claude 3.5 Sonnet failed to answer, many of which belonged to the hard category, including 109 from History and 48 from Drama and Theatre. 
Qwen2.5 and LLaMA-3.1 answered 220 and 190 of these missed questions, respectively.
These findings suggest that while closed models such as GPT-4o and Claude perform reasonably well in understanding general question structure and intent, they often fall short in capturing localized cultural nuances.
This suggests the necessity of developing models that are better attuned to the cultural and linguistic context of Sinhala.

\section{Conclusion}

We introduce SinhalaMMLU, the first multi-task language understanding benchmark designed to evaluate LLM capabilities in Sinhala, a low-resource language. This benchmark provides a structured evaluation based on the Sri Lankan educational curriculum. 
While closed-source models demonstrate higher levels of accuracy overall, both open and closed models struggle to achieve comparable results on culturally embedded concepts and language-specific features. 
This paved the need for future research on culturally aware and low-resource LLMs.
We believe that SinhalaMMLU will empower researchers to effectively evaluate and design Sinhala LLMs.

\section{Limitations}

Although we believe SinhalaMMLU will significantly contribute to Sinhala NLP and the design of next multilingual LLMs, it does have some limitations.

\paragraph{Text-based question only} SinhalaMMLU is limited to text-based multiple-choice questions, enabling standardized evaluation but excluding multimodal and open-ended tasks. Extending the benchmark to include images, tables, diagrams, audio and essay-style questions remains a valuable direction for future work.

\paragraph{Mathematics Not Included} We specifically exclude mathematics questions because the original exam papers do not include multiple-choice formats for this subject, focusing instead on problem-solving.
Additionally, Maths questions are already well covered by existing English math reasoning benchmarks.

\paragraph{Educational Scope}
The dataset primarily targets the secondary and collegiate levels of the Sri Lankan education system. Expanding the benchmark to include primary-level content, professional-level content, open-ended questions, and tasks that assess generative capabilities would provide a more comprehensive evaluation of LLMs in Sinhala.

\paragraph{Human evaluation}
Human evaluation was not conducted in this study due to practical limitations.
Evaluating subject-specific questions requires age-appropriate annotators with domain expertise, ideally during exam periods, which is logistically challenging. As an alternative, we provide official O-Level and A-Level pass rates as a reference\footnote{\url{https://www.gazette.lk/2025/05/gce-ordinary-level-in-sri-lanka.html}}.
However, we note that these statistics are drawn from official results for Sri Lanka's G.C.E. Ordinary Level and Advanced Level exams, and reflect only the overall pass rates of student cohorts for each year. While they are not directly comparable to our benchmark, since we focus solely on multiple-choice questions whereas the national exams include structured and essay-type questions, they still provide a meaningful reference point for understanding model performance in real-world educational settings.

\paragraph{Domain Imbalance and Sampling Bias}
Due to the curriculum structure and limited access to Sinhala exam papers with marking schemes, subjects in domains like humanities and social sciences are overrepresented. This imbalance may affect subject diversity and model performance comparisons.

\paragraph{Prompting Strategies and Evaluation Framework}
We did not experiment with chain-of-thought prompting or vision-language models, which may yield different results compared to our current approach.
While {lm-evaluation-harness} \cite{eval-harness} is used for evaluation recently, we adopted a custom evaluation pipeline aligned with prior work \cite{poh-etal-2024-malaymmlu, koto-etal-2024-arabicmmlu, li-etal-2024-cmmlu}.

\section{Ethical Statement}

The SinhalaMMLU dataset used in our study is collected from publicly available web resources, which are published by the government. 
In compliance with the Sri Lankan Copyright Act Number 36 year 2003\footnote{\url{https://www.gov.lk/elaws/wordpress/wp-content/uploads/2015/03/IntellectualPropertyActNo.36of2003Sectionsr.pdf}}, specifically section 11, the fair use of a work, including such use by reproduction in copies or by any other means specified by that section, for purposes such as criticism, comment, news reporting, teaching (including multiple copies for classroom use), scholarship or research, shall not be an infringement of copyright.
Our dataset is intended solely for research and educational purposes and complies with these fair use provisions.
The SinhalaMMLU benchmark will be released under an appropriate open license to support research and educational use.

We are committed to ensuring that the data collection and evaluation processes for SinhalaMMLU are conducted with the highest standards of transparency and fairness.
To support this, we adopted a crowd-sourcing approach for the annotation process, welcoming contributions from volunteers for data collection.
Individuals who provide significant contributions will be offered co-authorship, in accordance with the authorship guidelines set by the ACL, available at \href{https://www.aclweb.org/adminwiki/index.php/Authorship_Changes_Policy_for_ACL_Conference_Papers.}{Authorship Changes Policy for ACL Conference Papers.}
The annotators who contributed to the human evaluation will be acknowledged in the acknowledgments section after the review process, in accordance with anonymity guidelines.

Note that in this work, we used an AI assistant tool, Copilot, for coding support.
Additionally, our dataset may not fully reflect real-world exams, which often include multimodal and essay-style questions. This limitation should be considered when generalizing our findings to broader educational or practical applications.

\section*{Acknowledgments}
We would like to thank our annotators, Kavindu Akalanka Wickramasinghe and Dhanushi nagodage, for their valuable contributions to the human evaluation of STEM subjects.

\bibliography{custom}

\clearpage
\appendix

\section{Data Distribution}
\label{sec:app:subjects}

In this section, we provide detailed descriptions of the subjects categorized under each domain in Table~\ref{tab:subjects_by_domain}, and present the number of questions per subject across difficulty levels in Table~\ref{tab:Sinhalmmlu_Subject_distribution}. 
Questions for the ``Easy'' and ``Medium'' levels were collected from official examination papers with answer schemes, published between 2017 and 2023. While most ``Hard'' level questions were sourced from 2012 to 2023, in some subjects it was necessary to extend the search to the early 2000s in order to obtain appropriate questions and their corresponding answer schemes.

\begin{table}[ht]
\centering
\resizebox{\columnwidth}{!}{%
\begin{tabular}{c l c c c}
\toprule
\textbf{No.} & \textbf{Subject} & \textbf{Easy} & \textbf{Medium} & \textbf{Hard} \\
\midrule
1 & History & 154 & 121 & - \\
2 & Drama and Theatre & 131 & 140 & 143 \\
3 & Dancing & 111 & 171  & - \\
4 & Eastern Music & 146 & 113 & - \\
5 & Arts & 103 & 133 & - \\
6 & Buddhism & 165 & 120 & 147 \\
7 & Catholicism & 135 & 119 & - \\
8 & Christianity & 133 & 118 & 150 \\
9 & Islam & 111 & 120 & 119 \\
10 & Citizenship Education & 126 & 156 & - \\
11 & Health and Physical Science & 133 & 105 & - \\
12 & Geography & 120 & 157 & 127 \\
13 & Science & 167 & 112 & - \\
14 & Sinhala Language and literature & 158 & 120 & 119 \\
15 & Business and Accounting Studies & - & 122 & 109 \\
16 & Entrepreneurship Studies & - & 117 & - \\
17 & Home Economics & - & 150 & 140 \\
18 & Communication and Media Studies & - & 119 & 112 \\
19 & Agriculture and Food Technology & - & 136 & 158 \\
20 & Design and Construction Technology & -  & 117 & - \\
21 & Economics & - & - & 129 \\
22 & Biosystems Technology & - & - & 106 \\
23 & Buddhist Civilization & - & - & 119 \\
24 & Political Science & - & - & 160 \\
25 & Physics & - & - & 107 \\
26 & Chemistry & - & - & 116 \\
27 & Biology & - & - & 127 \\
28 & Oriental Music & - & - & 117 \\
29 & History of Sri Lanka & - & - & 147 \\
30 & Dancing Indigenous & - & - & 132 \\
\bottomrule
\end{tabular}%
}
\caption{Subject-wise counts of questions by  each difficulty level in the SinhalaMMLU dataset.}
\label{tab:Sinhalmmlu_Subject_distribution}
\end{table}

\paragraph{Educational Levels}

Table \ref{tab:sinhala_curriculum} summarizes the educational levels in Sri Lanka.

\begin{table}[htbp]
\centering
\resizebox{\columnwidth}{!}{%
\small
\begin{tabular}{@{}lllp{6.5cm}@{}}
\toprule
\textbf{Level} & \textbf{Grades} & \textbf{Age Range} & \textbf{Description} \\
\midrule
Primary (Excluded) & 1–5 & 5–9 & Focuses on foundational learning (e.g., drawing, writing, filling blanks); excluded from SinhalaMMLU due to unsuitability for MCQs. Ends with the optional Grade 5 Scholarship Exam.\\
\midrule
Junior Secondary & 6–9 & 10–13 & Covers core subjects; first phase of secondary education. Includes nine subjects as first language, English, mathematics, science and technology, social studies, life skills, religion, aesthetics, health, and physical education. A second language is also introduced. \\ \hdashline
Senior Secondary \\ (O-Level) & 10–11 & 14–16 & Prepares students for GCE O-Level exams with six main and three additional subjects. \\
\midrule
Collegiate \\ (A-Level) & 12–13 & 17–19/20 & Includes five academic streams: Arts, Commerce, Biological Science, Physical Science, and Technology. Students select three core subjects + General English + Common General Test. Qualification for university admission. \\
\bottomrule
\end{tabular}%
}
\caption{Structure of the Sri Lankan national education curriculum relevant to SinhalaMMLU.}%

\label{tab:sinhala_curriculum}
\end{table}

\subsection{Examples}

\begin{figure}[htbp]
\centering{
\includegraphics[width=\linewidth]{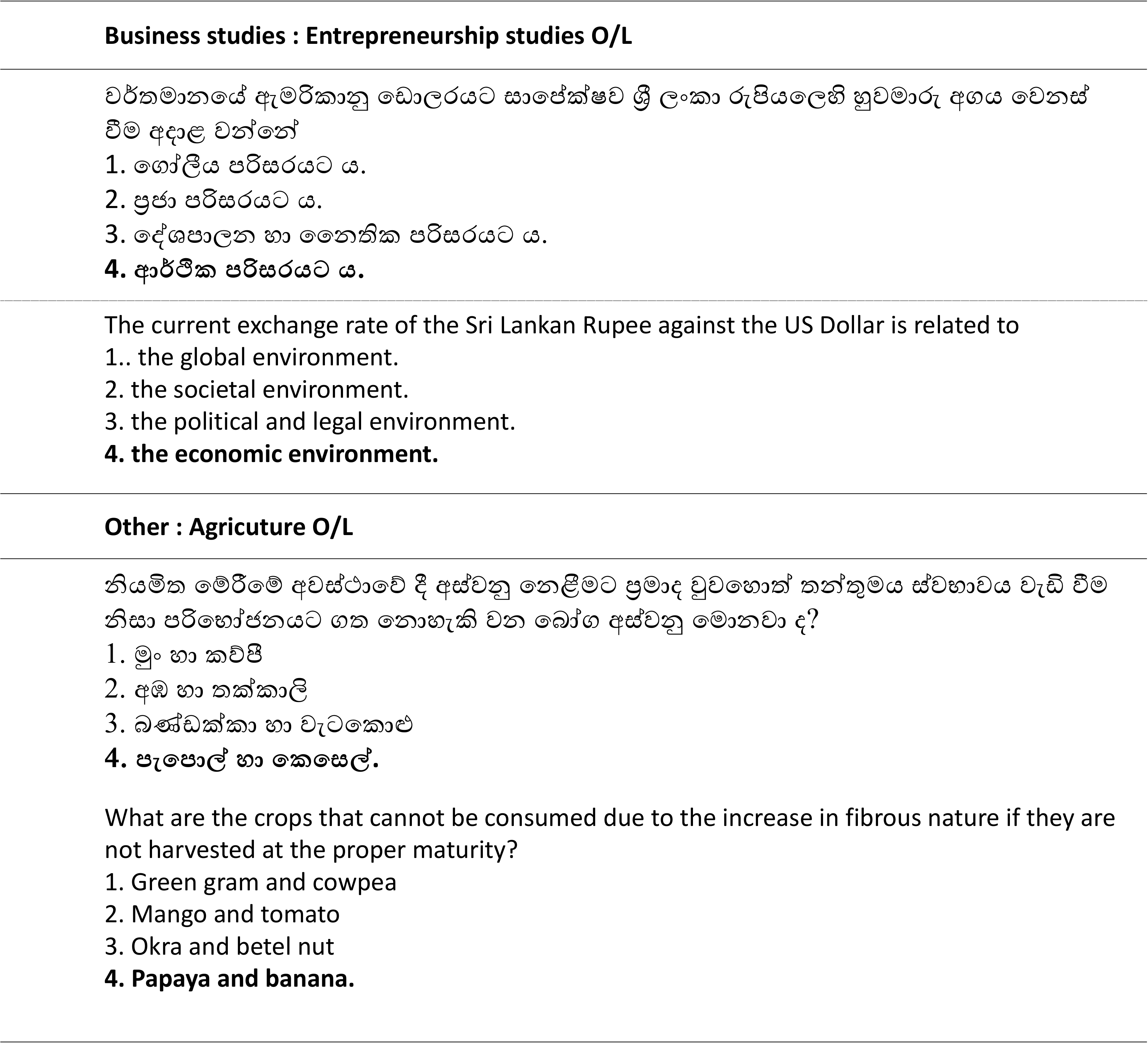}}
\caption{Example questions from the Business and Other domains.}
\label{ex_business}
\end{figure}

\begin{figure}[htbp]
\centering{
\includegraphics[width=\linewidth]{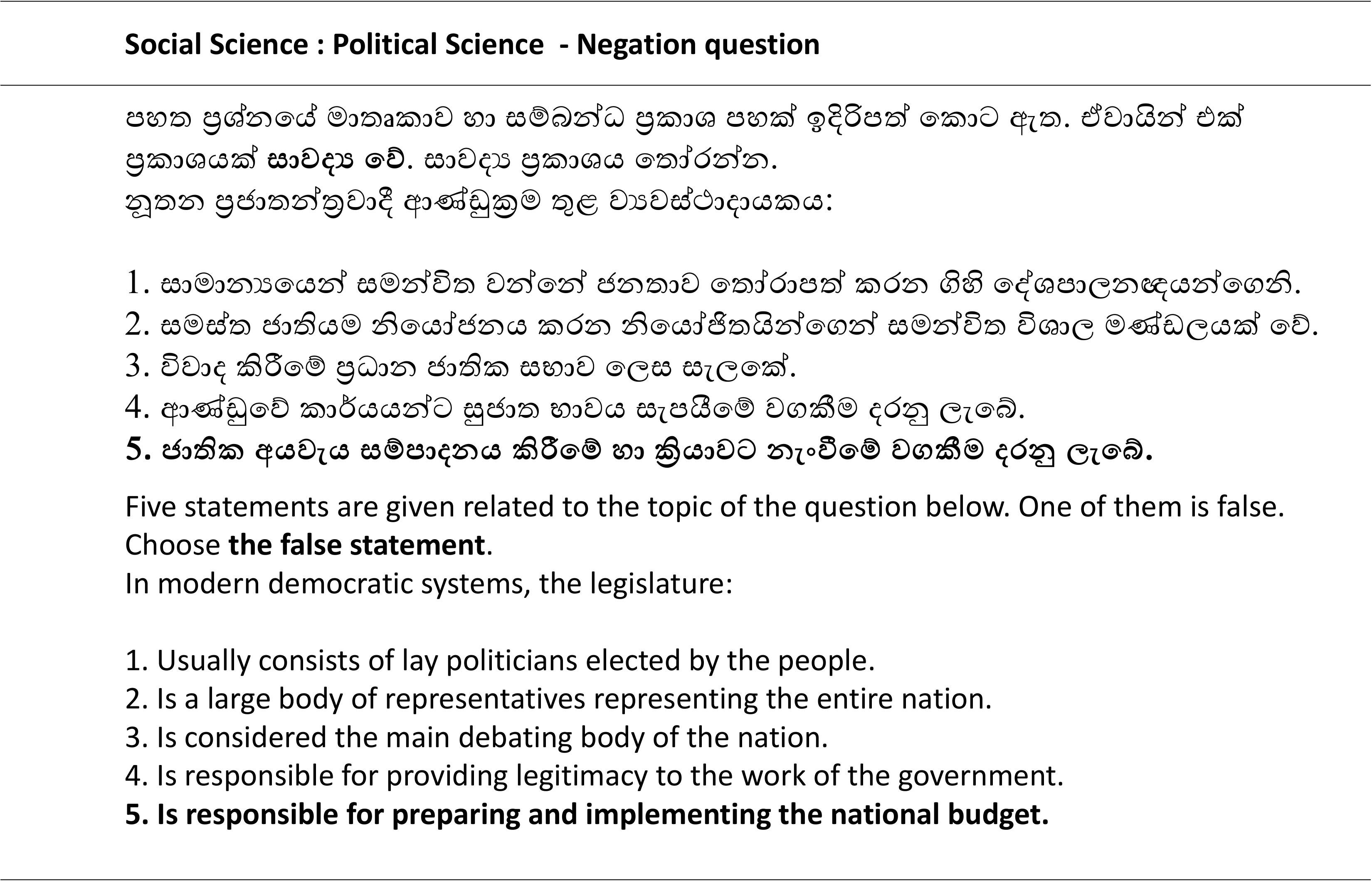}}
\caption{Example of negation question.}
\label{ex_negation}
\end{figure}

\begin{figure}[htbp]
\centering{
\includegraphics[width=\linewidth]{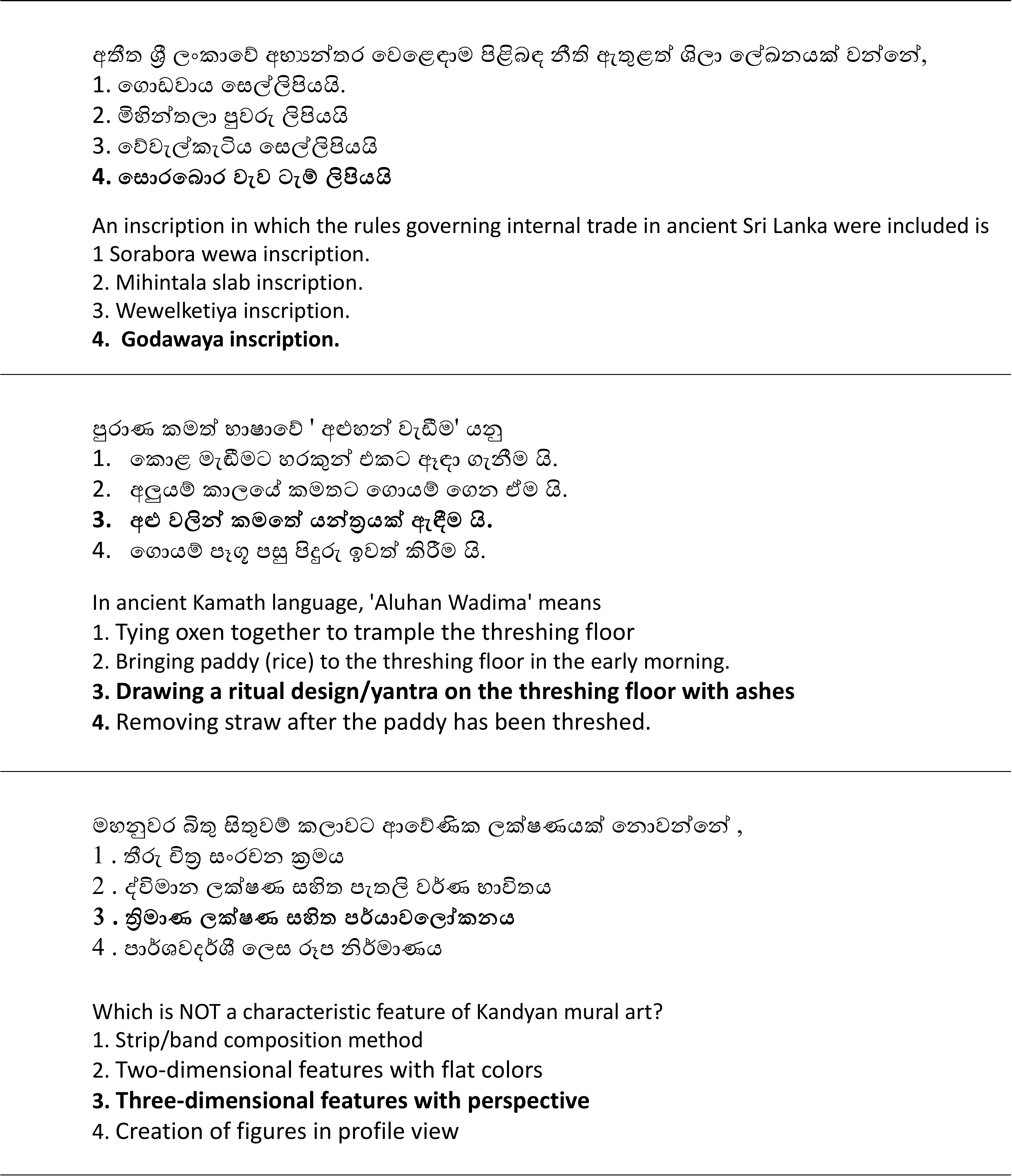}}
\caption{Examples of cultural questions from history, language, and arts.}
\label{ex_cul}
\end{figure}

Figure \ref{ex_business} illustrates example questions from the remaining domains, Business and Other, while Figure \ref{ex_negation} depicts a negation example. Examples of culturally relevant questions are provided in Figure \ref{ex_cul}.

\section{Experimental Setup}
\label{sec:Model Details}
All model generations were conducted on a single NVIDIA RTX A6000 Ada GPU. 
To optimize memory efficiency, we applied 4-bit quantization during inference.
Experiments with GPT-4o were run using {top\_p} = 0 and {temperature} = 0.5.

\paragraph{Model details} The open-source model and closed model artifacts used in our experiments are listed in Table \ref{tab:model-sources} and are accessible via the Hugging Face Hub except the GPT-4o and Claude.

\begin{table*}[tbp]
\centering
\small
\resizebox{\textwidth}{!}{%

\begin{tabular}{@{}lccccccccccccccccccccc@{}}
\toprule
\multirow{2}{*}{\textbf{Model}} & \multicolumn{5}{c}{\textbf{Easy}} & \multicolumn{7}{c}{\textbf{Medium}} & \multicolumn{7}{c}{\textbf{Hard}} \\
\cmidrule(lr){2-6} \cmidrule(lr){7-13} \cmidrule(lr){14-22}
 & Hum & Lang & SocSci & STEM & Avg & Hum & Lang & SocSci & STEM & Bus & Other & Avg & Hum & Lang & SocSci & STEM & Bus & Other & Avg \\
\midrule

\textsc{claude-3-5-sonnet} & 71.69 & 68.39 & 82.43 & 78.66 & 74.18 & 68.09 & 70.94 & 76.77 & 79.82 & 74.25 & 66.67 & 70.43 & 57.94 & 45.69 & 72.24 & 47.21 & 71.98 & 64.48 & 60.12 \\
\textsc{claude-3-haiku} & 45.87 & 46.45 & 62.16 & 50.00 & 49.54 & 42.46 & 48.72 & 52.08 & 44.95 & 50.64 & 39.80 & 44.65 & 34.06 & 35.34 & 33.10 & 25.22 & 37.50 & 39.48 & 34.22 \\
\textsc{gpt-4o-2024-08-06} & 68.42 & 50.32 & 74.59 & 77.44 & 68.94 & 62.59 & 58.97 & 71.39 & 78.90 & 67.38 & 62.55 & 65.00 & 54.33 & 44.83 & 67.26 & 46.63 & 67.67 & 60.32 & 56.71 \\ \hdashline

\textsc{Qwen2.5-7B} & 26.68 & 25.16 & 25.68 & 18.90 & 25.66 & 22.16 & 20.51 & 27.14 & 23.85 & 24.46 & 23.33 & 23.42 & 17.51 & 18.10 & 23.49 & 16.72 & 15.95 & 20.04 & 18.46 \\
\textsc{Qwen2.5-7B-chat} & 30.03 & 28.39 & 35.41 & 28.66 & 30.85 & 28.99 & 23.93 & 27.63 & 35.78 & 27.04 & 27.06 & 28.25 & 24.74 & 21.55 & 19.93 & 26.10 & 27.16 & 20.24 & 23.56 \\

\textsc{Qwen2.5-32B} & 33.39 & 35.48 & 38.11 & 34.15 & 34.58 & 29.96 & 23.08 & 36.67 & 30.28 & 30.04 & 30.20 & 30.81 & 22.65 & 24.14 & 29.54 & 21.41 & 20.26 & 24.80 & 23.52 \\
\textsc{Qwen2.5-32B-chat} & 41.39 & 38.71 & 43.51 & 49.39 & 42.30 & 37.06 & 33.33 & 38.88 & 47.71 & 35.19 & 39.02 & 37.87 & 31.11 & 24.14 & 33.10 & 30.50 & 34.48 & 28.97 & 30.81 \\

\textsc{Qwen2.5-72B} & 40.71 & 33.55 & 42.97 & 46.34 & 41.06 & 35.64 & 32.48 & 44.50 & 51.38 & 32.62 & 36.67 & 37.55 & 27.02 & 18.97 & 31.67 & 29.62 & 25.00 & 32.34 & 28.40 \\
\textsc{Qwen2.5-72B-chat} & 43.29 & 38.06 & 51.08 & 51.22 & 45.11 & 41.40 & 47.86 & 48.41 & 60.55 & 45.06 & 40.20 & 43.77 & 34.35 & 31.03 & 35.23 & 38.12 & 37.50 & 37.50 & 35.72 \\ \hdashline
\textsc{Llama-3-8B} & 26.94 & 25.16 & 26.22 & 22.56 & 26.26 & 22.34 & 21.37 & 27.87 & 24.77 & 24.89 & 23.53 & 23.78 & 17.70 & 17.24 & 24.20 & 16.42 & 16.81 & 19.44 & 18.50 \\
\textsc{Llama-3-8B-chat} & 27.97 & 24.52 & 27.84 & 22.56 & 27.17 & 22.34 & 20.51 & 28.36 & 22.94 & 23.61 & 24.51 & 23.82 & 17.79 & 18.97 & 23.84 & 16.72 & 15.95 & 21.83 & 19.01 \\
\textsc{Llama-3-70B} & 26.25 & 27.10 & 24.59 & 29.88 & 26.31 & 22.61 & 24.79 & 27.38 & 25.69 & 23.61 & 24.31 & 24.06 & 17.70 & 17.24 & 22.78 & 17.89 & 15.95 & 19.84 & 18.53 \\
\textsc{Llama-3-70B-chat} & 29.95 & 23.23 & 28.92 & 28.66 & 29.07 & 28.28 & 22.22 & 25.92 & 25.69 & 20.60 & 26.86 & 26.50 & 22.84 & 14.66 & 23.49 & 20.82 & 20.69 & 18.06 & 21.11 \\
\textsc{Llama-3.1-8B} & 26.33 & 25.16 & 28.92 & 20.73 & 26.26 & 22.78 & 26.50 & 29.83 & 26.61 & 27.47 & 24.12 & 24.98 & 19.22 & 18.10 & 21.00 & 16.72 & 19.40 & 22.62 & 19.72 \\
\textsc{Llama-3.1-8B-chat} & 28.83 & 29.68 & 34.05 & 23.17 & 29.44 & 26.68 & 21.37 & 31.05 & 22.02 & 24.46 & 26.86 & 26.78 & 19.89 & 24.14 & 19.57 & 21.99 & 17.67 & 23.02 & 20.75 \\
\textsc{Llama-3.1-70B} & 26.16 & 26.45 & 26.76 & 23.78 & 26.09 & 21.99 & 23.08 & 28.36 & 21.10 & 23.61 & 23.33 & 23.46 & 17.60 & 18.10 & 23.13 & 17.01 & 15.95 & 19.84 & 18.46 \\
\textsc{Llama-3.1-70B-chat} & 30.12 & 30.97 & 32.16 & 28.05 & 30.42 & 27.13 & 27.35 & 24.45 & 21.10 & 33.48 & 29.02 & 27.41 & 24.17 & 20.69 & 19.93 & 22.58 & 19.83 & 21.43 & 22.38 \\
\textsc{Llama-3.2-1B} & 26.59 & 25.16 & 25.41 & 18.90 & 25.55 & 22.16 & 20.51 & 27.14 & 22.94 & 23.61 & 23.53 & 23.34 & 17.32 & 18.10 & 23.84 & 16.72 & 15.95 & 19.84 & 18.38 \\
\textsc{Llama-3.2-1B-chat} & 26.76 & 25.16 & 25.68 & 18.90 & 25.72 & 22.25 & 20.51 & 27.14 & 22.94 & 23.61 & 23.53 & 23.38 & 17.32 & 18.10 & 23.84 & 17.30 & 15.95 & 19.84 & 18.46 \\
\textsc{Llama-3.2-3B} & 26.59 & 27.74 & 25.68 & 18.90 & 25.82 & 22.16 & 20.51 & 26.89 & 22.94 & 23.18 & 23.53 & 23.26 & 17.32 & 18.10 & 23.49 & 16.72 & 15.95 & 19.84 & 18.34 \\
\textsc{Llama-3.2-3B-chat} & 26.68 & 24.52 & 25.68 & 19.51 & 25.66 & 22.25 & 21.37 & 26.89 & 22.02 & 22.75 & 23.53 & 23.26 & 17.41 & 18.10 & 23.84 & 17.01 & 15.95 & 19.84 & 18.46 \\

\textsc{Llama-3.3-70B-chat} & 28.83 & 26.45 & 26.76 & 29.27 & 28.25 & 23.85 & 21.37 & 30.32 & 22.94 & 29.18 & 25.69 & 25.62 & 20.55 & 21.55 & 23.49 & 20.23 & 22.41 & 20.04 & 20.95 \\ \hdashline

\textsc{Mistral-7B} & 26.59 & 25.16 & 25.41 & 18.90 & 25.55 & 22.16 & 20.51 & 27.14 & 22.94 & 23.61 & 23.53 & 23.34 & 17.32 & 18.10 & 23.84 & 16.72 & 15.95 & 18.91 & 18.14 \\
\textsc{Mistral-7B-chat} & 23.92 & 22.58 & 25.91 & 27.44 & 24.49 & 22.96 & 27.35 & 32.27 & 22.94 & 23.18 & 21.76 & 24.46 & 17.41 & 18.97 & 20.28 & 18.42 & 16.81 & 19.84 & 18.37 \\ \hdashline

\textsc{aya-expanse-8b} & 24.87 & 23.87 & 26.76 & 21.34 & 24.85 & 23.32 & 29.91 & 25.67 & 26.61 & 23.18 & 23.33 & 24.14 & 19.89 & 18.97 & 22.06 & 16.72 & 17.67 & 20.04 & 19.49 \\

\textsc{aya-expanse-32b} & 22.20 & 21.94 & 35.14 & 28.05 & 25.28 & 27.39 & 33.33 & 28.61 & 33.03 & 26.61 & 26.67 & 27.89 & 22.26 & 20.69 & 25.27 & 21.11 & 20.69 & 22.62 & 22.30 \\

\bottomrule
\end{tabular}%
}
\caption{ Zero-shot model accuracy across six domains for each difficulty level (easy, medium, and hard.}
\label{zero-shot_all}
\end{table*}

\begin{table}[htbp]
\centering
\resizebox{\columnwidth}{!}{%
\begin{tabular}{lr}
\toprule
\textbf{Model} & \textbf{Source } \\
\midrule
\textsc{GPT-4o} & \texttt{gpt-4o-2024-08-06 } \\
\textsc{claude 3.5 sonnet} &\texttt{claude-3-5-sonnet-20241022} \\ 
\textsc{claude-3-haiku}  & \texttt{claude-3-haiku-20240307} \\ \hdashline
\textsc{aya-expanse-32b} & \texttt{CohereForAI/aya-32b} \\
\textsc{aya-expanse-8b} &\texttt{CohereForAI/aya-8b} \\ \hdashline
\textsc{Llama-3.2-1B-chat} & \texttt{meta-llama/Llama-3-1B-instruct} \\
\textsc{Llama-3.2-3B-chat} & \texttt{meta-llama/Llama-3-3B-instruct} \\
\textsc{Llama-3.3-70B-chat} & \texttt{meta-llama/Llama-3-70B-instruct} \\
\textsc{Llama-3.2-1B} & \texttt{meta-llama/Llama-3-1B} \\
\textsc{Llama-3.2-3B} & \texttt{meta-llama/Llama-3-3B} \\
\textsc{Llama-3-70B-chat} & \texttt{meta-llama/Meta-Llama-3-70B-Instruct} \\
\textsc{Llama-3-8B-chat} & \texttt{meta-llama/Meta-Llama-3-8B-Instruct} \\
\textsc{Llama-3.1-70B-chat} & \texttt{meta-llama/Meta-Llama-3.1-70B-Instruct} \\
\textsc{Llama-3.1-8B-chat} & \texttt{meta-llama/Meta-Llama-3.1-8B-Instruct} \\
\textsc{Llama-3-70B} & \texttt{meta-llama/Meta-Llama-3-70B} \\
\textsc{Llama-3-8B} & \texttt{meta-llama/Meta-Llama-3-8B} \\
\textsc{Llama-3.1-70B} & \texttt{meta-llama/Meta-Llama-3.1-70B} \\
\textsc{Llama-3.1-8B} & \texttt{meta-llama/Meta-Llama-3.1-8B} \\  \hdashline
\textsc{Mistral-7B-v0.3} & \texttt{mistralai/Mistral-7B-v0.3} \\
\textsc{Mistral-7B-chat-v0.3} & \texttt{mistralai/Mistral-7B-Instruct-v0.3} \\ \hdashline
\textsc{Qwen2.5-32B} & \texttt{Qwen/Qwen2.5-32B} \\
\textsc{Qwen2.5-32B-chat} & \texttt{Qwen/Qwen2.5-32B-Instruct} \\
\textsc{Qwen2.5-72B} & \texttt{Qwen/Qwen2.5-72B} \\
\textsc{Qwen2.5-72B-chat} & \texttt{Qwen/Qwen2.5-72B-Instruct} \\
\textsc{Qwen2.5-7B} & \texttt{Qwen/Qwen2.5-7B} \\
\textsc{Qwen2.5-7B-chat} & \texttt{Qwen/Qwen2.5-7B-Instruct} \\

\bottomrule
\end{tabular}%
}
\caption{Lists of the LLMs we used in this study and their corresponding Hugging Face IDs except for GPT-4o and Claude.}
\label{tab:model-sources}
\end{table}

\subsection{Prompt}
\label{sec:prompt}

\begin{figure}[htbp]
\centering{
\includegraphics[width=0.9\linewidth]{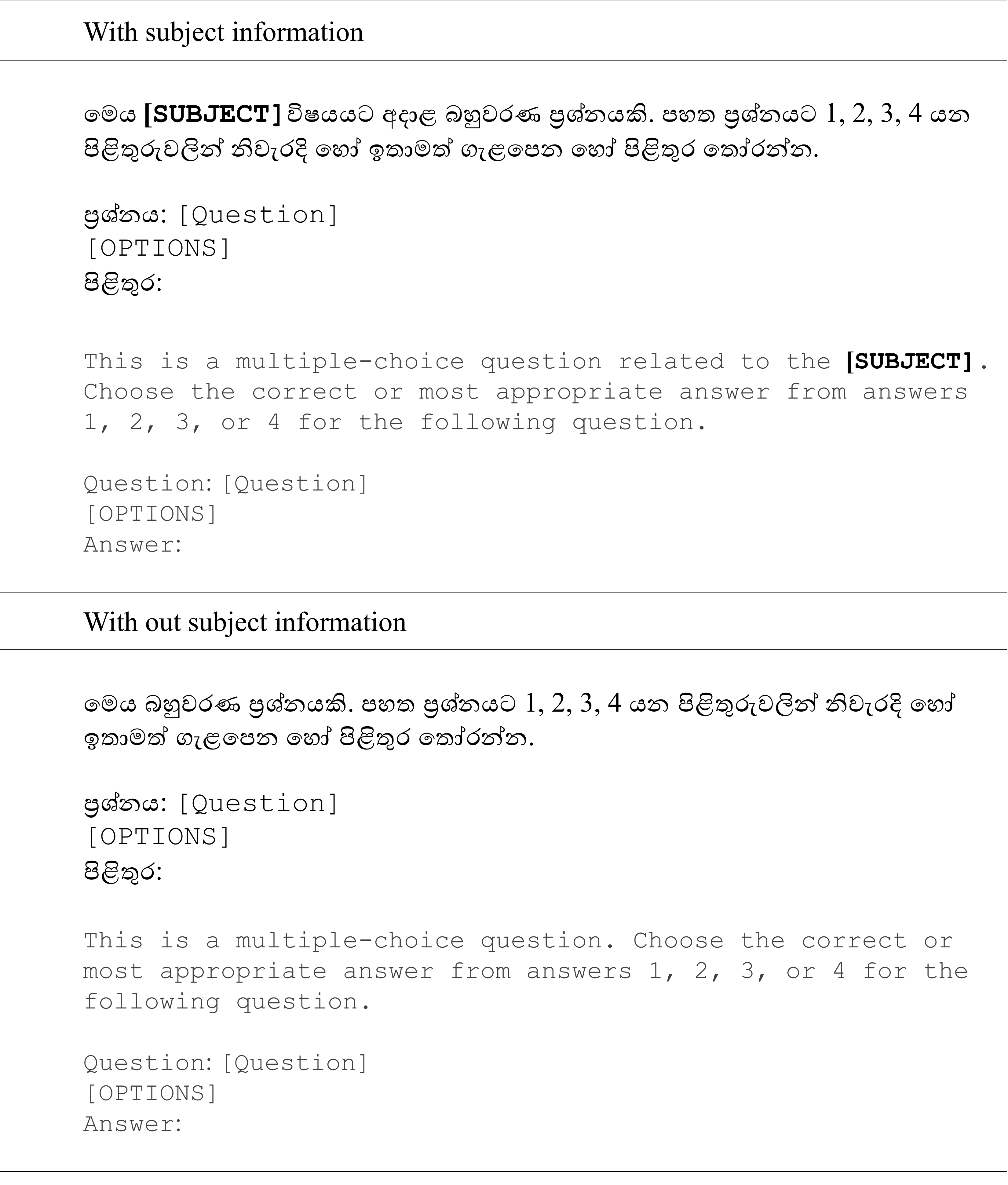}}
\caption{Prompt template used in the task.}
\label{prompt}
\end{figure}

\begin{figure*}[htbp]
\centering{
\includegraphics[scale=0.66]{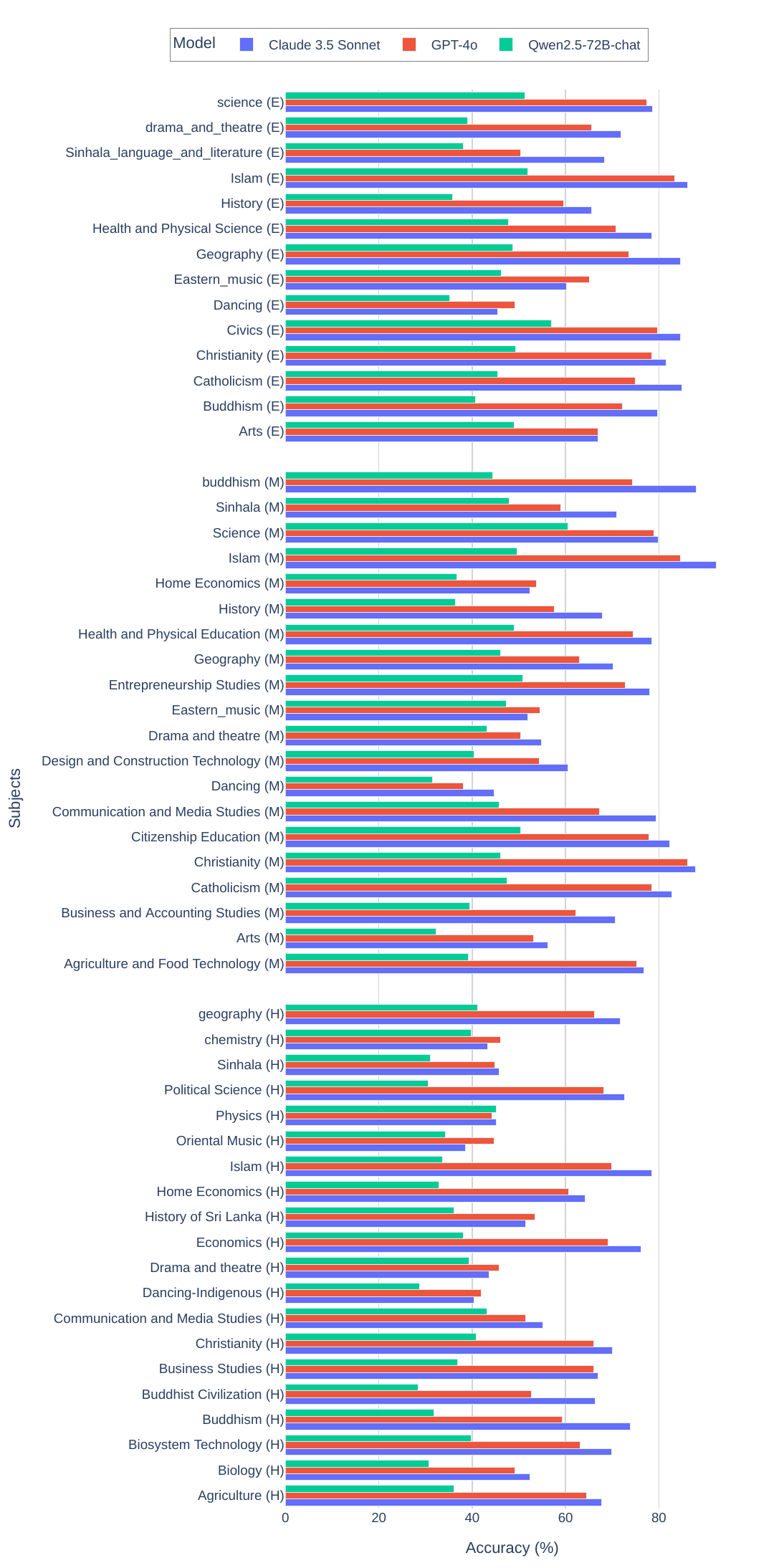}
\caption{Performance (\% accuracy) breakdown across the 30 subjects. ``E'', ``M'', and ``H'' indicate each difficulty level: easy, medium, and hard, respectively. }
\label{sub_acc}}
\end{figure*}

For the SinhalaMMLU evaluation, we used two distinct prompt templates, one that specifies the subject domain and another that omits this information. 
Both follow the instruction format proposed by \citet{hendryckstest2021}, with prompt instructions given in the same language as the question.
We observed that including the subject name generally improves performance across models. Therefore, for all subsequent experiments, we report results using prompts that include subject information.
For evaluation, we use the following prompting formats in \ref{prompt}. 
For 3-shot evaluation, the identical format is repeated.
The fewshot examples were gathered from random questions for each subject.
For closed models, we additionally use a system prompt to specify the expected answer format. 
All experiments involving closed models were conducted using the official APIs provided by OpenAI and Anthropic.

\section{Additional Results and Discussions}

\label{sec:prompt_compare}
\begin{figure}[htbp]
\centering{
\includegraphics[width=\linewidth]{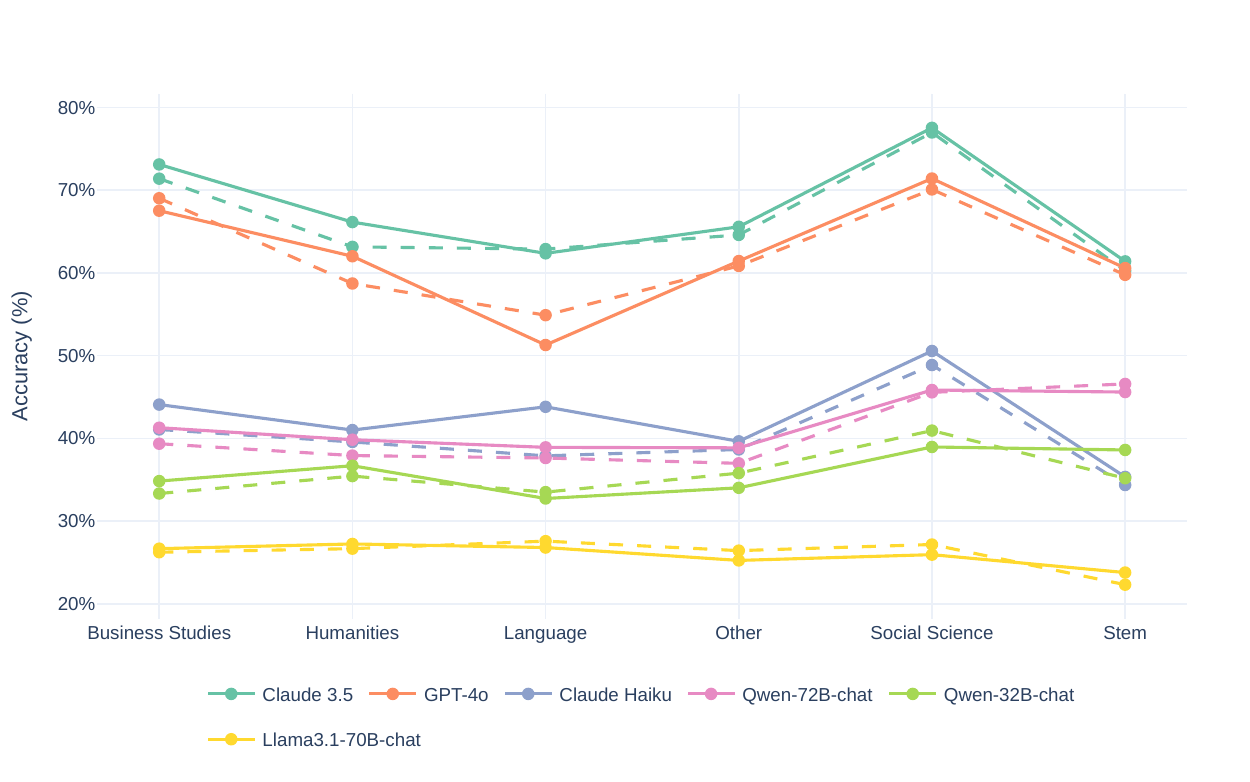}}
\caption{Accuracy variation of LLMs across 6 domains under two prompting strategies with and without including the subject in the prompt. Solid lines indicate prompts including the subject; dashed lines indicate prompts without it. Each model is represented by a consistent color.}
\label{fig:prompt_sub}
\end{figure}


\paragraph{Does including the subject affect performance?}
To evaluate the impact of the subject, we compared two prompt variants: with and without the subject. 
As shown in Figure \ref{fig:prompt_sub}, Claude 3.5 Sonnet's accuracy improves across all domains when the subject is included, while GPT-4o performs worse on Sinhala language tasks. This suggests GPT may have weaker prior knowledge of Sinhala grammar or context compared to Claude, making it more reliant on explicit prompting.

\subsection{Results by difficulty}
\label{sec:diff_results}
Table \ref{zero-shot_all} presents the detailed zero-shot results across six domains for each difficulty level.

\begin{figure}[htbp]
\centering{
\includegraphics[width=\linewidth]{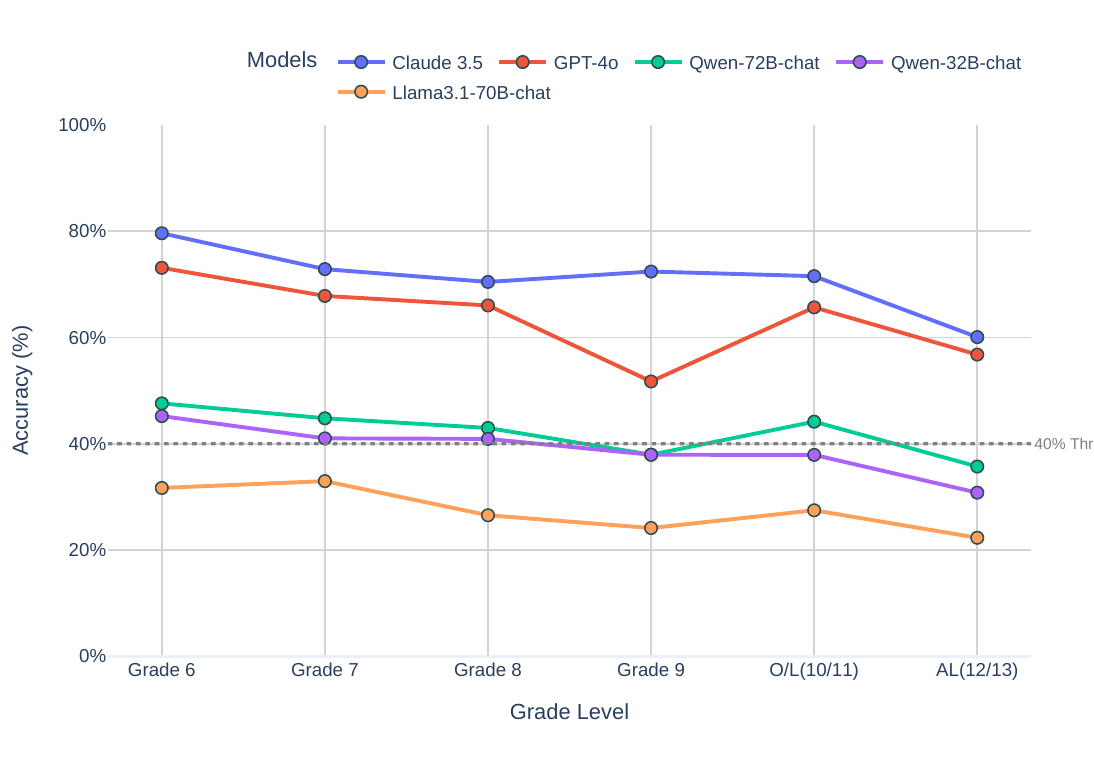}}
\caption{LLM accuracy by grade level with a 40\% threshold line. Grades with no questions are excluded.}
\label{grade_acc}
\end{figure}

\subsection{ Results By Subject}
\label{sec:Subject_results}

Subject-wise analysis shows that models tend to achieve their highest scores in subjects that require less analytical reasoning, such as Buddhism, Christianity, Islam Citizenship Education, and Health, where questions are more fact-based and rely on descriptive knowledge.  
In contrast, lower scores are observed in culturally grounded subjects like Sinhala language, History, and Drama and Theatre, which require deeper cultural and contextual understanding. A more detailed breakdown of subject-wise performance is provided in the Figure~\ref{sub_acc}.

\subsection{Results by grade}

As illustrated in Figure \ref{grade_acc}, SinhalaMMLU includes metadata on educational grade levels, enabling a more fine-grained evaluation of LLM performance from an educational perspective.
In the Sri Lankan education system, a minimum score of 40\% is typically required to pass. 
However, the current benchmark evaluates only the multiple-choice component of examination papers, which represent 40–50\% of the total marks, the remainder consisting of structured and essay-type questions.
Interestingly, performance drops at grades 9, with models performing better on GCE O-Level exam questions, this is basically due to that grade 9 only contained questions in arts which skewed the results. 
Most open source models barely reach this threshold, suggesting limited readiness for real exam scenarios.

\subsection{Few-shot vs zero-shot}
\label{sec: app_fewshot}
Table \ref{tab:category-accuracy} presents the detailed few-shot results across six domains, while Table \ref{tab:results_fewshot_difficulty_level} compares few-shot and zero-shot performance for each difficulty level.

\begin{table*}[htbp]
\resizebox{\textwidth}{!}{%
\centering
\begin{tabular}{lccccccc}
\toprule
\textbf{Model} & \textbf{Humanities} & \textbf{Language} & \textbf{Social Science} & \textbf{Stem} & \textbf{Business Studies} & \textbf{Other} & \textbf{Average} \\
\midrule
\textsc{Llama-3-8B} & 24.84 & 26.55 & 30.19 & 23.55 & 20.65 & 21.33 & 24.92 \\
\textsc{Llama-3-8B-chat} & 22.45 & 19.59 & 25.44 & 19.22 & 19.78 & 20.77 & 22.03 \\
\textsc{Llama-3-70B} & 22.61 & 21.39 & 25.40 & 22.16 & 20.65 & 21.77 & 22.59 \\
\textsc{Llama-3-70B-chat} & 23.30 & 23.16 & 24.05 & 27.20 & 19.14 & 23.51 & 23.35 \\ 
\textsc{Llama-3.1-8B} & 28.28 & 27.32 & 34.43 & 27.15 & 26.88 & 25.19 & 28.60 \\
\textsc{Llama-3.1-8B-chat} & 23.47 & 23.71 & 27.90 & 23.29 & 21.94 & 23.20 & 23.99 \\
\textsc{Llama-3.1-70B} & 23.29 & 22.68 & 25.19 & 21.16 & 19.78 & 21.44 & 22.90 \\
\textsc{Llama-3.1-70B-chat} & 25.16 & 26.47 & 29.15 & 23.43 & 20.86 & 23.20 & 25.05 \\
\textsc{Llama-3.2-1B} & 23.29 & 20.10 & 25.09 & 22.16 & 23.87 & 22.98 & 23.30 \\
\textsc{Llama-3.2-1B-chat} & 22.18 & 21.65 & 25.64 & 18.40 & 19.78 & 22.21 & 22.17 \\
\textsc{Llama-3.2-3B} & 23.86 & 20.88 & 25.00 & 21.76 & 22.80 & 22.65 & 23.47 \\
\textsc{Llama-3.2-3B-chat} & 22.69 & 21.39 & 25.05 & 21.17 & 20.22 & 19.89 & 22.28 \\
\textsc{Llama-3.3-70B-chat} & 26.09 & 24.63 & 30.31 & 24.18 & 20.00 & 21.74 & 25.45 \\
\textsc{Mistral-7B-chat} & 22.87 & 22.68 & 25.83 & 21.56 & 24.09 & 21.66 & 23.13 \\
\textsc{Qwen2.5-32B} & 28.94 & 28.87 & 32.71 & 30.62 & 25.16 & 31.05 & 29.68 \\
\textsc{Qwen2.5-32B-chat} & 37.29 & 35.05 & 41.45 & 39.74 & 32.47 & 36.24 & 37.54 \\
\textsc{Qwen2.5-72B} & 35.05 & 27.58 & 43.87 & 36.16 & 29.89 & 30.94 & 35.20 \\
\textsc{Qwen2.5-72B-chat} & 40.59 & 36.86 & 46.70 & 44.30 & 39.57 & 37.90 & 41.24 \\
\textsc{Qwen2.5-7B} & 22.18 & 21.65 & 25.64 & 18.40 & 19.78 & 22.10 & 22.15 \\
\textsc{Qwen2.5-7B-chat} & 28.23 & 26.55 & 31.04 & 26.06 & 25.38 & 28.40 & 28.18 \\
\textsc{aya-expanse-32b} & 24.91 & 25.74 & 30.26 & 27.20 & 21.72 & 22.40 & 25.21 \\
\textsc{aya-expanse-8b} & 22.30 & 21.65 & 25.64 & 17.43 & 20.22 & 21.77 & 22.11 \\
\bottomrule
\end{tabular}%
}
\caption{Few-shot accuracy per each domain.}
\label{tab:category-accuracy}
\end{table*}

\begin{table*}[htbp]
\centering
\begin{tabular}{lcccccc}
\toprule
\multirow{2}{*}{Model} & \multicolumn{2}{c}{Easy} & \multicolumn{2}{c}{Medium} & \multicolumn{2}{c}{Hard} \\
\cmidrule(lr){2-3} \cmidrule(lr){4-5} \cmidrule(lr){6-7}
 & Zero-Shot & 3-Shot & Zero-Shot & 3-Shot & Zero-Shot & 3-Shot \\
\midrule
\textsc{Llama-3.2-1B} & 25.55 & 25.77 & 23.34 & 24.74 & 18.38 & 19.76 \\
\textsc{Llama-3.2-1B-chat} & 25.72 & 25.54 & 23.38 & 23.34 & 18.46 & 18.42 \\
\textsc{Llama-3.2-3B} & 25.82 & 26.63 & 23.26 & 24.50 & 18.34 & 19.80 \\
\textsc{Llama-3.2-3B-chat} & 25.66 & 26.76 & 23.26 & 22.98 & 18.46 & 18.21 \\
\textsc{Llama-3.3-70B-chat} & 28.25 & 29.63 & 25.62 & 25.14 & 20.95 & 20.64 \\
\textsc{Llama-3-70B} & 26.31 & 27.11 & 24.06 & 24.98 & 18.53 & 18.41 \\
\textsc{Llama-3-70B-chat} & 29.07 & 25.26 & 26.50 & 25.36 & 21.11 & 19.47 \\
\textsc{Llama-3-8B} & 26.26 & 29.66 & 23.78 & 26.50 & 18.50 & 19.41 \\
\textsc{Llama-3-8B-chat} & 27.17 & 24.38 & 23.82 & 23.46 & 19.01 & 18.79 \\
\textsc{Llama-3.1-70B} & 26.09 & 25.72 & 23.46 & 24.66 & 18.46 & 18.71 \\
\textsc{Llama-3.1-70B-chat} & 30.42 & 28.80 & 27.41 & 26.14 & 22.38 & 19.81 \\
\textsc{Llama-3.1-8B} & 26.26 & 34.31 & 24.98 & 30.65 & 19.72 & 21.80 \\
\textsc{Llama-3.1-8B-chat} & 29.44 & 28.08 & 26.78 & 25.14 & 20.75 & 19.74 \\
\textsc{Mistral-7B-chat} & 24.49 & 25.15 & 24.46 & 25.46 & 18.37 & 19.02 \\
\textsc{Qwen2.5-32B} & 34.58 & 36.26 & 30.81 & 27.77 & 23.52 & 26.74 \\
\textsc{Qwen2.5-32B-chat} & 42.30 & 41.57 & 37.87 & 40.42 & 30.81 & 31.54 \\
\textsc{Qwen2.5-72B}& 41.06 & 49.05 & 37.55 & 25.54 & 28.40 & 34.60 \\
\textsc{Qwen2.5-72B-chat} & 45.11 & 46.25 & 43.77 & 43.85 & 35.72 & 34.69 \\
\textsc{Qwen2.5-7B} & 25.66 & 25.54 & 23.42 & 23.30 & 18.46 & 18.42 \\
\textsc{Qwen2.5-7B-chat} & 30.85 & 31.01 & 28.25 & 30.41 & 23.56 & 23.76 \\
\textsc{aya-expanse-32b} & 25.28 & 28.75 & 27.89 & 25.54 & 22.30 & 19.60 \\
\textsc{aya-expanse-8b} & 24.85 & 25.59 & 24.14 & 23.50 & 19.49 & 18.05 \\
\bottomrule
\end{tabular}%
\caption{Model Performance Comparison (Zero-Shot vs 3-Shot).}
\label{tab:results_fewshot_difficulty_level}

\end{table*}

\end{document}